\documentclass{article}

\usepackage{PRIMEarxiv}

\errorcontextlines\maxdimen

\usepackage[utf8]{inputenc} 
\usepackage[T1]{fontenc}    
\usepackage{url}            
\usepackage{booktabs}       
\usepackage{nicefrac}       
\usepackage{microtype}      
\usepackage{fancyhdr}       
\usepackage{graphicx}
\usepackage{xcolor}
\usepackage{threeparttable}
\usepackage{subcaption}

\usepackage{lineno,hyperref}
\hypersetup{colorlinks, citecolor=blue, linkcolor=red, urlcolor=green}
\usepackage{amsmath,amsfonts,amssymb}
\usepackage{graphics}
\usepackage{bm}
\usepackage{multirow}

\usepackage{algorithm}
\usepackage{algpseudocode}

\usepackage{mathtools}
\usepackage{siunitx}
\DeclarePairedDelimiterXPP\BigOSI[2]%
  {\mathcal{O}}{(}{)}{}%
  {\SI{#1}{#2}}

\title{Generative adversarial wavelet neural operator: Application to  fault detection and isolation of multivariate time series data}

\author{Jyoti Rani\\
  Department of Chemical Engineering\\
  Indian Institute of Technology Delhi\\
  Hauz Khas, 110016, India\\
  \texttt{jyoti.rani@chemical.iitd.ac.in} \\
  \And
  Tapas Tripura\\
  Department of Applied Mechanics\\
  Indian Institute of Technology Delhi\\
  Hauz Khas, 110016, India\\
  \texttt{tapas.t@am.iitd.ac.in} \\
  \And
  Hariprasad Kodamana \\
  Department of Chemical Engineering \\
  Yardi School of Artificial Intelligence (ScAI)\\
  Indian Institute of Technology Delhi\\
  Hauz Khas, 110016, India\\
  \texttt{kodamana@iitd.ac.in} \\
  \And
  Souvik Chakraborty \\
  Department of Applied Mechanics\\
  Yardi School of Artificial Intelligence (ScAI)\\
  Indian Institute of Technology Delhi\\
  Hauz Khas, 110016, India\\
  \texttt{souvik@am.iitd.ac.in} \\
}

\begin{document}
\maketitle

\begin{abstract}
Fault detection and isolation in complex systems are critical to ensure reliable and efficient operation. However, traditional fault detection methods often struggle with issues such as nonlinearity and multivariate characteristics of the time series variables. 
This article proposes a generative adversarial wavelet neural operator (GAWNO) as a novel unsupervised deep learning approach for fault detection and isolation of multivariate time series processes.
The GAWNO combines the strengths of wavelet neural operators and generative adversarial networks (GANs) to effectively capture both the temporal distributions and the spatial dependencies among different variables of an underlying system. The approach of fault detection and isolation using GAWNO consists of two main stages. In the first stage, the GAWNO is trained on a dataset of normal operating conditions to learn the underlying data distribution. 
In the second stage, a reconstruction error-based threshold approach using the trained GAWNO is employed to detect and isolate faults based on the discrepancy values. We validate the proposed approach using the Tennessee Eastman Process (TEP) dataset and Avedore wastewater treatment plant (WWTP) and $\mathrm{N}_2 \mathrm{O}$ emissions named as $\mathrm{WWTP} \mathrm{N}_2 \mathrm{O}$ datasets. Overall, we showcase that the idea of harnessing the power of wavelet analysis, neural operators, and generative models in a single framework to detect and isolate faults has shown promising results compared to various well-established baselines in the literature.
\end{abstract}


\section{Introduction}
In an industrial setting,  fault detection and isolation (FDI) is a crucial activity because it can help to reduce inefficiencies, prevent catastrophic failures, and prevent unnecessary shutdowns \cite{venkatasubramanian2003review1,venkatasubramanian2003review2,venkatasubramanian2003review3,wang2022integrated}. The quantity of data processed and archived through industrial operations has grown substantially in recent years due to the adoption of Industry 4.0 and subsequent developments. As a result, data-driven frameworks have  gained considerable interest in fault detection and isolation 
\cite{qin2012survey,weese2016statistical,reis2017industrial,raveendran2018process,yu2012fault}.
Due to their usefulness in modeling highly dimensional data, multivariate statistical techniques like principal component analysis (PCA) \cite{chiang2000fault,kodamana2017mixtures,de2022comparing} and partial least-squares (PLS) \cite{ge2017review} have traditionally been employed for process monitoring. However, these techniques only consider cross-correlations between variables and ignore the temporal dynamics of the time series data. Thus, they are only appropriate for tracking the static variations in the data. 
Methods like slow feature analysis (SFA) \cite{song2022slow}, dynamic principal component analysis (DPCA) \cite{rato2013fault}, and dynamic inner canonical correlation analysis (DiCCA) \cite{dong2019efficient} have been suggested to incorporate dynamics in multidimensional data to address this issue. DPCA extends PCA by incorporating time-varying dynamics into the principal components. While high-order autoregressive mathematical models are utilized in DiCCA to extract dynamic latent variables to describe the dynamics in latent space \cite{dong2019efficient}, SFA can be used to progressively extract slow-changing features by emphasizing first-order dynamics \cite{song2022slow}. The ability of these techniques to simulate system dynamics enables fault-finding algorithms to track changes in process data that are both dynamic and static \cite{zhao2018full,dong2019efficient,rani2023fault,fang2018novel,fang2018novel2}. 

In recent years, the use of neural networks for fault detection has also seen acceptance among the scientific and industrial communities due to their ability to learn complicated nonlinear characteristics from vast amounts of process data. One such fault detection framework is based on the autoencoder neural networks, which compresses the data into latent dimensions and reconstructs them while learning the representation of the process data \cite{pumsirirat2018credit,goswami2023fault,mo2023deep}. By learning the underlying patterns in the data and identifying deviations from these patterns, the autoencoders are able to identify the presence of a fault. 
As the autoencoder model lacks a robust regularisation, so it often tends to be overfitted. Variational autoencoders (VAE) are among other alternatives that aim to model the latent space with a regularized Gaussian probability distribution  \cite{an2015variational,GOSWAMI20231885}.  In VAE, the reconstruction error between the generated and actual samples is used to measure the extent of the anomaly. 
Attention-based models have also been proposed for fault detection, which uses a transformer-based architecture to generate and compare deviations in the time series data \cite{mohammadi2020transformer}. 

A  deep generative learning framework, namely,  the generative adversarial networks (GAN), aims at learning data distribution \cite{goodfellow2014generative}. GAN is composed of two neural networks: a generator network that generates new data and a discriminator network that differentiates between actual and generated data. 
Architectures like Recurrent Neural Network (RNN) \cite{su2019robust,peng2024application} and its primary variant, the Long Short-Term Memory (LSTM) \cite{park2018multimodal,osarogiagbon2020new} have also been used most frequently as models for the generators and discriminators to capture the implicit relationship of the time series data. Some of the examples are the Recurrent Conditional GAN (RCGAN) \cite{esteban2017real}, the Time GAN (TimeGAN) \cite{yoon2019time}, and the Conditional Sig-Wasserstein GAN (SigCWGAN) \cite{ni2020conditional}. It finds a wide range of potential applications as a generative model for images, text, and time series data \cite{mi2023wgan}. 
However, GANs also find their application in the domain of fault detection and isolation \cite{liu2023anomaly}. 
Fault detection using generative networks involves learning the underlying patterns from typical operating data and detecting deviations from these patterns without explicit fault modeling. A few of the popular GAN frameworks for fault detection are MAD-GAN \cite{li2019mad}, TAno-GAN \cite{bashar2020tanogan}, and Tad-GAN \cite{geiger2020tadgan} methods. 

A limitation associated with the above data-driven neural networks (NNs) is their tendency to exhibit poor generalization beyond the data used for training. The generalization of trained networks is considered an efficient and effective approach, whereby the networks are expected to predict outputs with maximum accuracy even for unseen inputs. In order to tackle these challenges, there has been a recent development of neural operators \cite{li2020neural,lu2021learning,kovachki2021neural}. Neural operators learn about complex nonlinear operators by passing global integral operators through nonlinear local activation functions \cite{chen1995universal}. By approximating the operators between function spaces, the neural operators generalize beyond the sample distribution. 
One of the initial works in neural operators is the deep operator network (DeepONet) \cite{lu2021learning,garg2022assessment}, which was developed predominantly for solving partial differential equations (PDEs). 
Another immediate development in the neural operator domain includes the Fourier neural operator (FNO), where the network parameters are learned in Fourier space \cite{li2020fourier,RANI20231897}. The spectral decomposition of the input in FNO helps in learning only the key features of the input, resulting in robust learning of solution operators and generalization over unseen scenarios. 

In a recent study, the wavelet neural operator (WNO) is proposed \cite{tripura2023wavelet}, which uses a spatiotemporal decomposition of the signals as compared to the Fourier decomposition in FNO. Since the wavelet basis functions are both spatial and frequency localized \cite{boggess2015first,wirsing2020time}, the use of wavelet analysis offers helpful insight into both the characteristic frequency and characteristic space information of a signal.
As a result of the inherited ability of wavelets to capture the complex patterns in the data, WNO is able to generalize beyond training data, especially in complex boundary conditions. 
Since WNO was first proposed, it has experienced remarkable advancements and specialized developments, for e.g., WNO for medical elastography \cite{tripura2023elastography}, physics-informed wavelet neural operator (PIWNO) \cite{tripura2023physics}, randomized prior wavelet neural operator (RPWNO) \cite{garg2023randomized}, probabilistic wavelet neural operator auto-encoder (PWNOAE) \cite{rani2023fault} and WNO for foundation modeling \cite{tripura2023foundational}. 
Although the PWNOAE proposed in \cite{rani2023faulta} was able to learn the first and second-order statistics of the infinite-dimensional multivariate input time series data, it has shown that it has constraints to learn the exact underlying multivariate probability distributions. To address this shortcoming, we propose the generative adversarial wavelet neural operator (GAWNO) model. The proposed GAWNO employs the concepts of the classical generative adversarial network (GAN) to learn the underlying data distributions; however, it models the underlying generators and discriminators as the WNO. Further, unlike the WNO, it adopts a U-Net architecture for the generator WNO and discriminator WNO modules. By integrating the WNO with concepts of GAN, the proposed GAWNO is able to learn complex multivariate probability distributions. 
The ability to learn correct data distribution is then utilized for fault detection and isolation in multivariate data, wherein the first stage, the proposed GAWNO, is trained using data from normal operating conditions. In the second stage, for any new unseen data, the reconstruction error is computed and investigated against a threshold discrepancy value. Case studies for fault detection and isolation in the Tennessee Eastman Process (TEP) data and wastewater treatment data for various fault criteria showcase promising results in terms of notable efficacy and success in detecting and isolating faults in the time series data compared to various well-established baselines in the literature.

\section{Background}

\subsection{Generative Adversarial Networks}
\begin{figure}[ht!]
    \centering
    \includegraphics[width=\textwidth]{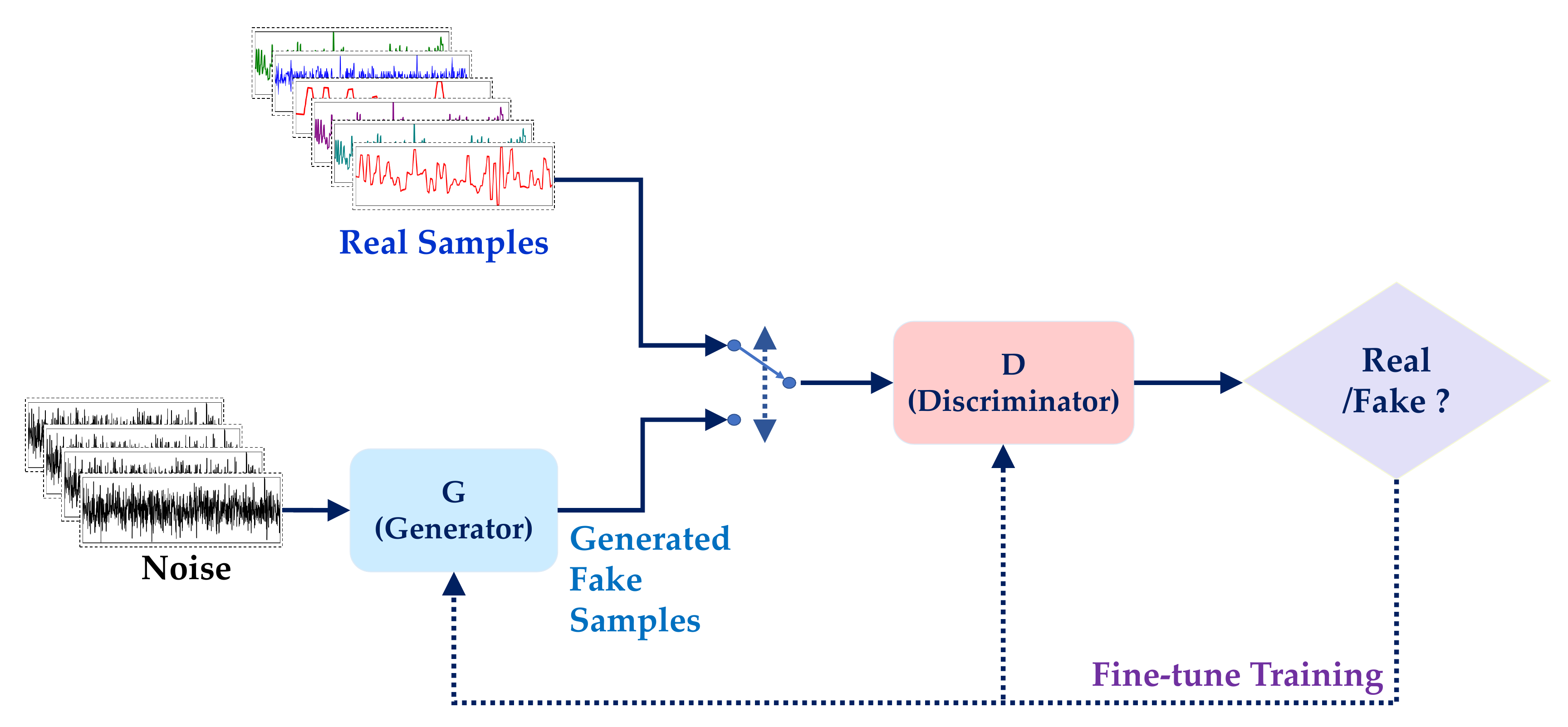}
    \caption{Generative Adversarial Network architecture (GAN). The GAN architecture for time series data is characterized by its two integral components: the generator and the discriminator. The input to the generator is typically a random noise vector sampled from a latent space. Through an iterative learning process, the generator learns the underlying temporal dependencies and patterns within the training data, aiming to generate time series samples that resemble the statistical characteristics of the original dataset. In contrast, the discriminator acts as an adversarial critic, distinguishing between real-time series data from the training set and synthetic data generated by the generator. As the generator aims to generate increasingly realistic samples, the discriminator concurrently strives to become more proficient at detecting fake data, creating a dynamic interplay that drives both components to improve over time.}
    \label{fig:GAN}
\end{figure}

Generative Adversarial Network (GAN) is a type of neural network architecture that is used to learn the underlying probability distribution of a dataset through adversarial training of a generator network $G(\cdot)$ and a discriminator network $D(\cdot)$, as shown in Figure \ref{fig:GAN}. The generator network $G(\cdot)$ takes random noise $\bm{z}$ as input, which is typically sampled from a distribution like standard normal or a uniform distribution. The generator network then transforms this noise into synthetic data samples $\hat{\bm{x}} = G(\bm{z})$ that should resemble the real data samples $\bm{x}$. 
The discriminator network $D(\cdot)$ takes in both real data samples $\bm{x}$ and synthetic data samples $\hat{\bm{x}}$ generated by the generator network $G(\cdot)$. The discriminator network then outputs a probability indicating whether the input data sample is real, represented by $D(\bm{y})$, where $\bm{y}=\{\bm{x},\hat{\bm{x}}\}$ is the input data sample to the discriminator. The discriminator network $D(\cdot)$ is trained to maximize the probability of correctly identifying real data samples and minimizing the probability of falsely identifying synthetic data samples as real.
The networks are trained iteratively in a minimax game framework to improve the quality of the generated synthetic data samples $\hat{\bm{x}}$ by obtaining the following objective function:
\begin{equation}
    \min _G \max _D V(D, G)=\mathbb{E}_{\bm{x} \sim p_{\text {data }}}[\log D(\bm{x})]+\mathbb{E}_{\bm{z} \sim p_{\bm{z}}(z)}[\log (1-D(G(\bm{z})))].
\end{equation}
The first term of the objective function encourages the discriminator network to correctly identify real data samples, while the second term encourages the generator network to generate synthetic data samples that can convince the discriminator network to think they are real.
Through repeated training iterations, the generator network learns to generate synthetic data samples whose probability distribution $p(\hat{\bm{x}})$ closely matches the real data distribution $p(\bm{x})$. On the other hand, the discriminator network becomes better at distinguishing between real and synthetic data samples as described in the following steps,
\begin{itemize}
    \item Training of Discriminator, when network parameters of Generator are fixed,
    \begin{equation}
       \mathcal{L}_D = \max _D V\left(D, G^*\right)=\mathbb{E}_{\bm{x} \sim p_{\text {data }}}[\log D(\bm{x})]+\mathbb{E}_{\bm{z} \sim p_{\bm{z}}(z)}\left[\log \left(1-D\left(G^*(\bm{z})\right)\right)\right],
    \end{equation}
    where $G^*(\cdot)$ indicates the generator $G(\cdot)$ with fixed network parameters nad $\mathcal{L}_D$ indicates the discriminator loss. We can see that optimizing $D$ is to maximize the difference between the \textit{real} sample $\bm{x}$, and \textit{fake} sample $\hat{\bm{x}} = G^*(\bm{z})$.
    \item Training of Generator when network parameters of Discriminator are fixed,
    \begin{equation}
       \mathcal{L}_G = \min _G V\left(D^*, G\right)=\mathbb{E}_{\bm{x} \sim p_{\text {data }}}\left[\log D^*(\bm{x})\right]+\mathbb{E}_{\bm{z} \sim p_{\bm{z}}(z)}\left[\log \left(1-D^*(G(\bm{z}))\right)\right].
    \end{equation}
    Above $\mathcal{L}_G$ is the loss function for the training Generator $G(\cdot)$. The loss function value is computed from a Discriminator with fixed network parameters, whose inputs are the \textit{real} samples $\bm{x}$ and \textit{fake} samples $\hat{\bm{x}}$.
    \item After computing the discriminator and generator losses, we update the weights of both networks using gradient descent or other optimization techniques. For the discriminator, the weights are updated to minimize \(\mathcal{L}_D\):
    \begin{equation}
         \operatorname{Update}_D = -\nabla_{\bm{\theta}_D} \mathcal{L}_D, 
    \end{equation}
    where $\bm{\theta}_D$ represents the discriminator's weights.
        \item Similarly, for the generator, the weights are updated to minimize \(Loss_G\):
    \begin{equation}
         \operatorname{Update}_G = -\nabla_{\bm{\theta}_G} \mathcal{L}_G, 
    \end{equation}
    where $\bm{\theta}_G$ represents the generator's weights. The gradients ($\nabla$) are calculated through back-propagation, and the weights are adjusted to move in the direction that reduces the respective losses.
\end{itemize}
The training process is iterative, with multiple rounds of updates for both the discriminator and the generator. As the discriminator improves at distinguishing real and generated data, the generator adapts to produce more convincing data, leading to a dynamic interplay between the two networks.
Through this adversarial training, the basic GAN learns to generate data that closely matches the real data distribution, while the discriminator becomes more challenged in differentiating between the two data types. This equilibrium leads to the generation of high-quality, realistic data by the generator.

\subsection{Wavelet Neural Operator}
Wavelet Neural Operator (WNO) is a data-driven framework for discovering mappings between two infinite-dimensional function spaces. In contrast to conventional neural networks (fully connected and convolutional networks), which can only learn a single mapping between the input and output, the WNO learns the family of input-output mappings. In WNO, the learning of the family of input-output mappings takes place through three sequential steps, which include an uplifting transformation, a wavelet integral block, and a downlifting transformation.
To discuss further, we consider the input-output pair $\{ {z(t)}, {\hat{x}(t)} \}$, for ${z(t)} \in \mathcal{A}$ and ${\hat{x}}(t) \in \mathcal{U}$, where $\mathcal{A} \in \mathbb{R}^{d_{z}}$ and $\mathcal{U} \in \mathbb{R}^{d_{x}}$ are the input and output function spaces. In the first step, the input ${z(t)}$ goes through an uplifting transformation as below:
\begin{equation}
    {\rm{P}}: z(t) \in \mathbb{R}^{d_a} \mapsto v(t) \in \mathbb{R}^{d_v},  
\end{equation}
where the high-dimensional transformation can be either a linear layer \cite{tripura2023wavelet} or a $1\times 1$ convolution layer \cite{tripura2023elastography}.
The wavelet integral block takes the lifted output and performs certain recursive integral transformations that are analogous to the convolutional hidden layers, except the fact that in WNO, the convolutions are performed in wavelet space. For more details on the analogy between the wavelet integral block and classical operator theory, one can refer to \cite{tripura2023physics}.
The recursive iterations are denoted as $\mathcal{F}: {v_{j}}(t) \in \mathbb{R}^{d_v} \mapsto v_{j+1}(t)\in \mathbb{R}^{d_v}$ for $j=1,\ldots,J$, and defined as,
\begin{equation}\label{eq:updates}
    v_{j+1}(t) = \sigma \left( \left( \mathcal{K}(t, v_j(t);\phi\in \theta)\right)+ \ell{v_j}(t)\right); \; t>0,
\end{equation}
where $\sigma(\cdot)$ is a scalar-valued non-linear activation function, $\ell : \mathbb{R}^{d_v}\rightarrow \mathbb{R}^{d_v}$ is a linear transformation, $*$ is the convolution operator, and $\mathcal{K}$ denotes the wavelet integral operator parameterized by $\phi \in \theta$, where $\theta$ is the network parameter. 
In the above equation, the convolution between the kernel integral operator and the uplifted input $v_j(t)$ is defined as,
\begin{equation}\label{eq:convolve}
    \mathcal{K}(t, v_{j}(t); \phi \in \theta) \triangleq \int_{T} k_{\phi \in \theta } \left(t- \tau \right) v_{j}(\tau) \mathrm{d} \tau; \quad t \in [0,T],
\end{equation}
where $k(\cdot,\cdot)$ is the kernel of the neural network parameterized by $\phi \in \theta$.
Parameterization of the kernel $k(\cdot,\cdot)$ in the wavelet space is expressed as,
\begin{equation}\label{eq:weight_conv}
    \mathcal{K}(t,v_j(t);\phi \in \theta) = \mathcal{W}^{-1}(\mathcal{R}_\phi \cdot \mathcal{W}(v_j))(t),
\end{equation}
where $\mathcal{R}_\phi = \mathcal{W}(k_{\phi})$ with $\mathcal{W}(\cdot)$ and $\mathcal{W}^{-1}(\cdot)$ denoting the forward and inverse wavelet transform.
The forward and inverse wavelet transforms $\mathcal{W}(v_j)$ and $\mathcal{W}^{-1}(v_j)$ are defined as,
\begin{align}
    \mathcal{W}_{(\zeta,\tau)}(v_j(t)) &= {\int_{-{\infty}}^{\infty} {\zeta}^{-1/2}\psi {\left({\zeta}^{-1}\left({t-\tau}\right) \right)} v_j(t)dt}, \label{eq:cwt} 
    \\
    \mathcal{W}^{-1}_{(\zeta,\tau)}(v_j(\zeta,\tau)) &= {{c_{\psi}}^{-1}}{ \int_{{0}}^{\infty}\int_{-{\infty}}^{\infty}  \mathcal{W}_{(\zeta,\tau)}(v_j(t)) {{\psi_{(\zeta,\tau)}(t)}}{\zeta^{-1/2}}d\tau d\zeta }, \label{eq:icwt}
\end{align}
where $\psi$ represents the orthonormal mother wavelet which is scaled and translated using the scaling and shifting parameters $\tau$ and $\zeta$, respectively, to form a family of wavelet basis functions, i.e., $\psi_{(\zeta, \tau)} = \psi {\left({\zeta}^{-1}\left({t-\tau}\right) \right)}$. The term $c_{\Psi}$ is an admissible constant defined as :
\begin{equation}
    c_{\Psi} \triangleq { \int_{-{\infty}}^{\infty} {\psi(v)}{|v|}^{-1}dv }
\end{equation}
with $\psi(v)$ being the Fourier transform of $v(t)$.
At the conclusion of $J$ iterations, a downlifting transformation $\rm{Q}$ 
is performed to transform the lifted space to the output space ${\hat{x}(t)} \in \mathcal{U}$ as follows: 
\begin{equation}
    {\rm{Q}}: v_{J}(t) \in {\mathbb{R}^{d_v}} \rightarrow \hat{x}(t) \in {\mathbb{R}^{d_x}}
\end{equation} 
This transformation can again be a linear layer or a $1 \times 1$ convolutional layer.
\begin{figure}
    \centering
    \includegraphics[width=0.8\textwidth]{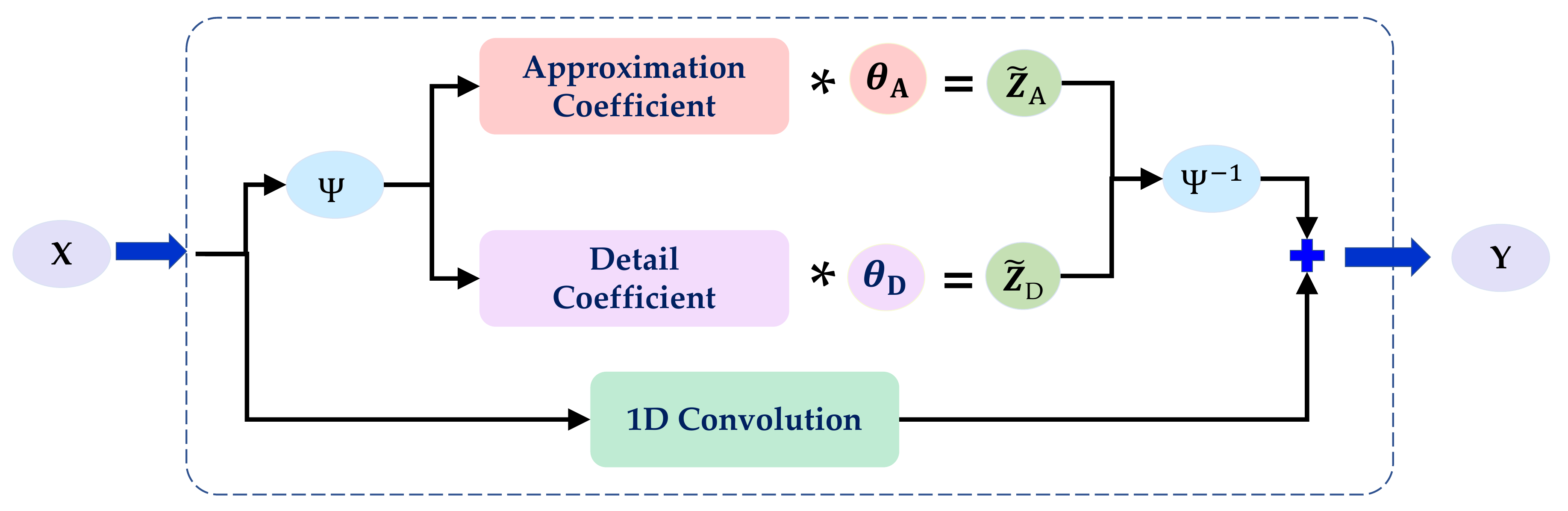}
    \caption{Schematic of the wavelet integral blocks. The input data is simultaneously passed through a wavelet filter and a 1-D CNN layer. In the wavelet filter, two levels of multi-resolution wavelet decomposition happen. The decomposed wavelet coefficients in the second level are used for kernel parameterization. Once the kernel convolution is done, the convolved data are transformed back to the time domain using inverse wavelet transform. In the CNN layer, a kernel size of one is used to preserve the shape of the data. Outputs of the wavelet parameterization and CNN are added to construct the output of the wavelet integral block.}
    \label{fig:integral_block}
\end{figure}


\section{Architecture of the proposed Generative Adversarial Wavelet Neural Operator}
The proposed Generative Adversarial Wavelet Neural Operator (GAWNO) extends the existing Generative Adversarial Networks (GAN) to an infinite-dimensional space to support the adversarial min-max game. The model consists of two main components: (a) a generator neural operator and (b) a discriminator neural function.
The architecture of GAWNO is built upon the concept of wavelet neural operator (WNO), which acts as mappings between two function spaces. Inspired by the U-net architecture, which is effective for finite-dimensional input and output, GAWNO incorporates WNO with increased depth, skip connections between layers, and information compression within reduced function spaces.
The generator neural operator takes a multivariate random noise variable as input and generates a multivariate sample from the desired probability distribution. On the other hand, the discriminator neural function includes a neural operator and an integral function. It processes either synthetic or real data as input and outputs scalar probabilities for the generated multivariate samples. 

\begin{figure}[t!]
    \centering
    \includegraphics[width=\textwidth]{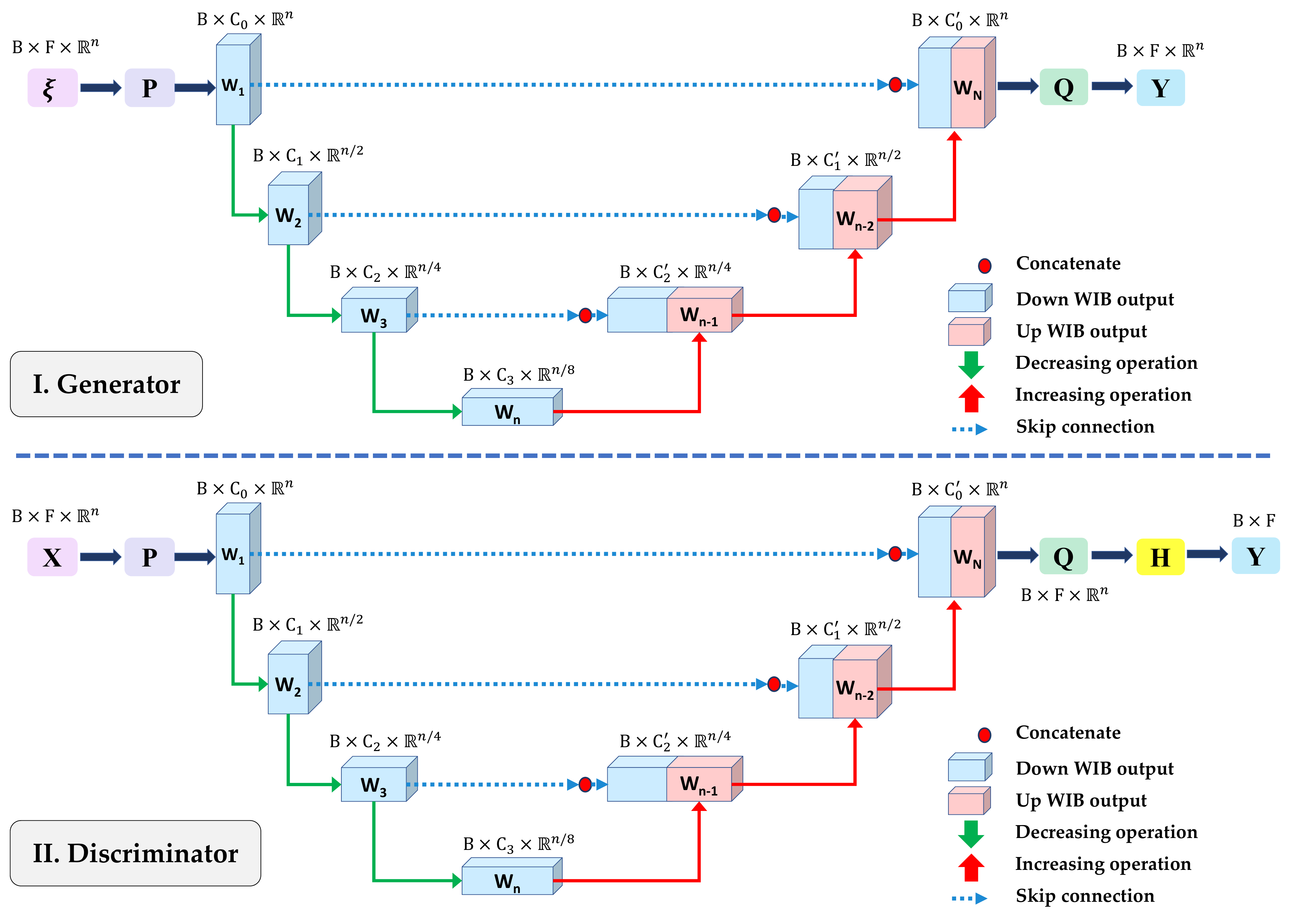}
    \caption{Generative adversarial Wavelet neural operator (GAWNO). (I) Generator: The input $\xi$ is random noise which first gets passed to a pointwise lifting operator parameterized with P. Then multiple layers of wavelet intergal blocks are applied, which are accompanied by a few skip connections. At last, the output Y is generated using a final pointwise projection layer parameterized with Q. (II) Discriminator: The input (real data and fake data) first gets passed to a pointwise lifting operator parameterized with P. Then multiple layers of wavelet intergal blocks are applied which are accompanied by a few skip connections. At last, the output Y is generated using a final pointwise projection layer parameterized with Q, followed by a linear integral functional layer H.}
    \label{fig:gener_discrem}
\end{figure}

In the first half of the operations in generator layers, each input function space is mapped to vector-valued function spaces with progressively smaller domains while increasing the number of output channels. 
In the later half of the generator layers, the domain of each intermediate function space is gradually expanded while reducing the dimension of the output channels. 
During the reconstruction phase in the later half, skip connections are employed to create vector-valued functions with higher co-dimension.
This process allows for the encoding of function spaces over smaller domains and facilitates direct information transfer across different grid sizes and scales. Further, by progressively reducing the domain of each input function space in the earlier layers, GAWNO becomes a memory-efficient architecture that enables deeper and more parameterized neural operators. The discriminator neural functional layer follows a similar architecture, with the integral functional layer implemented using a three-layer neural network.
Overall, GAWNO leverages the power of WNO, skip connections, and information encoding within reduced function spaces to create an effective and efficient architecture for generating desired probability distributions over function data. The GAWNO framework is portrayed in Fig. \ref{fig:GAN}, whose generator and discriminator neural operators are illustrated in Fig. \ref{fig:gener_discrem}.

\subsection{Generator neural operator}
The architecture of the generator neural operator $G(\cdot)$ is illustrated in Fig. \ref{fig:gener_discrem}. It takes random noises $\bm{z}(t) \in \mathbb{R}^{B \times F \times n} \sim \mathcal{N}(0, 1)$ as input and generates synthetic samples ${\hat{\bm{x}}}(t) \in \mathbb{R}^{B \times F \times n}$, where $B$ is the batch number, $F$ is the input feature number, and $n$ is the measurement length. 
Following the WNO architecture, the uplifting transformation ${\rm{P}}: \mathbb{R}^{B \times F \times n} \mapsto \mathbb{R}^{B \times {C_0} \times n}$ is applied to the input, where ${C_0} > F$ is the uplifted dimension. In our work, the transformation $\rm{P}$ is modeled as a linear neural network with a single hidden dimension. 
Similarly, in the end, the nonlinear solution transformation ${\rm{Q}}: \mathbb{R}^{B \times C_{o}^{\prime} \times n} \mapsto \mathbb{R}^{B \times F \times n}$, $C_{o}^{\prime} > C_{o}$, consisting of an activated linear neural network with two hidden layers is applied. 
The intermediate processes consist of the iterations defined in Eq. \eqref{eq:updates} organized in a U-Net type architecture. The U-Net architecture using skip connections between wavelet integral blocks enhances the co-dimensionality of the functions within the intermediate layers. However, to perform the parameterization of the wavelet integral operator $\mathcal{K}$ in the U-Net architecture, we introduce uplifting and downlifting wavelet integral blocks. The parameterization of the kernels $k^{up}_{\phi}$ and $k^{down}_{\phi}$ of the uplifting and downlifting wavelet integral operators $\mathcal{K}^{up}$ and $\mathcal{K}^{down}$ are represented as follows:
\begin{align}
    \mathcal{K}^{up}(t,v_j(t);\phi \in \theta) &= \mathcal{W}^{-1}_{up}(\mathcal{R}^{up}_\phi \cdot \mathcal{W}_{up}(v_j))(t), \\
    \mathcal{K}^{down}(t,v_j(t);\phi \in \theta) &= \mathcal{W}^{-1}_{down}(\mathcal{R}^{down}_\phi \cdot \mathcal{W}_{down}(v_j))(t),
\end{align}
where $\mathcal{R}^{up}_\phi = \mathcal{W}(k^{up}_{\phi})$, and $\mathcal{R}^{down}_\phi = \mathcal{W}(k^{down}_{\phi})$ with $\mathcal{W}_{up}$ and $\mathcal{W}_{down}$ being the forward uplifting and downlifting wavelet transforms, respectively. The details about the uplifting and downlifting wavelet transforms are discussed in section \ref{sec:updown}.
The U-Net architecture consists of four downlifting wavelet integral blocks (denoted in blue in Fig. \ref{fig:gener_discrem}) followed by four uplifting wavelet integral blocks (denoted in red in Fig. \ref{fig:gener_discrem}). In the downlifting wavelet integral blocks, the input channels are progressively increased while the measurement dimension is recursively compressed (half the dimension of the previous layer). Similarly, in the uplifting wavelet integral blocks, the lifted channels are progressively decreased, and the actual measurement dimension is recovered. The training of the generator focuses on updating and minimizing:
\begin{equation}
    \nabla_{\theta_G} \frac{1}{m} \sum_{i=1}^m \log \left(1-D\left(G\left(z^{(i)}\right)\right)\right),
\end{equation}
where $\theta_{G}$ is a hyperparameter of a multilayer perception that represents a differentiable function $G(\bm{z}; \theta_{G})$ which maps input noise z to data space.
\begin{algorithm}[ht!]
\caption{GAWNO (Generative Adversarial Wavelet Neural Operator) model Training }
\label{algo:gan-unet-wno}
\begin{algorithmic}[1]
    \Require{Normal time series training data: $\mathbf{x}_{\text{train}}$, epochs: $T$, loss function: $\mathcal{L}$, optimizer for generator: $O_G$, optimizer for discriminator: $O_D$.}
    \State Initialize network weights for Wavelet Neural Operator generator: $\bm{\theta}_G$, and Wavelet Neural Operator discriminator: $\bm{\theta}_D$.
    \For{$t \gets 1$ to $T$}
        \State Sample a batch of normal training time series data: $\bm{x} \sim \bm{x}_{\text{train}}$.
        \State Calculate discriminator loss on real data: $\mathcal{L}(D(\bm{x}), 1)$.
        \State Sample a batch of noise: $\bm{z} \sim \mathcal{N}(0,1)$.
        \State Generate synthetic samples: $\bm{x}_{\text{synth}} \gets G(\bm{z})$.
        \State Calculate discriminator loss on synthetic data: $\mathcal{L}(D(\bm{x}_{\text{synth}}), 0)$.
        \State Calculate total discriminator loss: $\mathcal{L}_D \gets \mathcal{L}(D(\bm{x}), 1) + \mathcal{L}(D(\bm{x}_{\text{synth}}), 0)$.
        \State Update discriminator weights: $\bm{\theta}_D \gets O_D(\nabla_{\theta_D} \mathcal{L}_D)$,
            $$
                \bm{\theta}_D \gets \underset{\bm{\theta}_D}{\text{arg\,min}} \left( -\frac{1}{2}\mathbb{E}_{\bm{x} \sim \mathbf{x}_{\text{train}}}\left[\log D(\bm{x})\right] -\frac{1}{2}\mathbb{E}_{\bm{z} \sim \mathcal{N}(0,1)}\left[\log\left(1 - D(\bm{x}_{\text{synth}})\right)\right] \right).
            $$
        \State Sample a batch of noise: $\bm{z} \sim \mathcal{N}(0,1)$.
        \State Generate synthetic samples: $\bm{x}_{\text{synth}} \gets G(\bm{z})$.
        \State Calculate generator loss: $\mathcal{L}_G \gets \mathcal{L}(D(\bm{x}_{\text{synth}}), 1)$.
        \State Update generator weights: $\bm{\theta}_G \gets O_G(\nabla_{\theta_G} \mathcal{L}_G)$,
            $$
                \bm{\theta}_G \gets \underset{\bm{\theta}_G}{\text{arg\,min}} \left( -\frac{1}{2}\mathbb{E}_{\bm{z} \sim \mathcal{N}(0,1)}\left[\log\left(1 - D(G(\bm{z}))\right)\right] \right).
            $$
    \EndFor
    \Ensure Trained generator weights $\bm{\theta}_G$ and discriminator weights $\bm{\theta}_D$.
\end{algorithmic}
\end{algorithm}

\subsection{Discriminator neural operator}
The architecture of the discriminator is similar to the generator, except it outputs the probabilities of the generated synthetic samples being false or true.
Unlike the random noise inputs to the generator, the discriminator takes the output of generator ${\hat{\bm{x}}}(t) \in \mathbb{R}^{B \times F \times n}$ and the corresponding actual samples $\bm{x}(t) \in \mathbb{R}^{B \times F \times n}$ as inputs and computes the probabilities of generated samples being fake. Discriminator first uplifts the input using the transformation $\rm{P}:\mathbb{R}^{B \times F \times n} \to \mathbb{R}^{B \times {C_0} \times n}$ and performs the iterations in Eq. \eqref{eq:updates} using uplifting and downlifting wavelet integral operators. A nonlinear transformation ${\rm{Q}}: \mathbb{R}^{B \times C_{o}^{\prime} \times n} \mapsto \mathbb{R}^{B \times F \times n}$, computes an intermediate function $\bm{h} \in \mathbb{R}^{B \times F \times n}$. Thereafter the output probabilities of the discriminator $\bm{r} \in \mathbb{R}^{B \times F}$ are computed as,
\begin{equation}
    \bm{r} = \int_T \kappa_d(t) \bm{h}(t) dt
\end{equation}
where the function $\kappa_d$ is parameterized as a 3-layered fully connected neural network. Noting that $\bm{h}$ is the output of the inner neural operator with $\hat{\bm{x}}$ as an input; therefore, $\bm{h}$ directly depends on $\hat{\bm{x}}$.
The function $k_d$ constitutes the integral functional $\int \kappa_d(x)$, which acts point-wise on its input function. This linear integral functional as the last layer is the direct generalization of the last layer of discriminators in GAN models to map a function to a number. By utilizing this discriminator architecture, the model aims to differentiate between real and generated samples, assigning a discriminating score to each input function u. While the discriminator is trained, it classifies both the real data and the fake data from the generator, it penalizes itself for misclassifying a real instance as fake or a fake instance (created by the generator) as real by maximizing the below function. The training of the discriminator focuses on updating and maximizing:
\begin{equation}
    \nabla_{\theta_G} \frac{1}{m} \sum_{i=1}^m\left[\log D\left(x^{(i)}\right)+\log \left(1-D\left(G\left(\hat{x}^{(i)}\right)\right)\right)\right]
\end{equation}
More details of the wavelet integral blocks are provided in later sections.

\subsection{Uplifting and Downlifting Wavelet Integral blocks}\label{sec:updown}
Wavelet integral blocks (WIB) are responsible for the processing of input function data at different wavelet scales and efficient parameterization of the kernel $\mathcal{R}_{\phi}$.
The process of continuous wavelet decomposition, as described in Eq. \eqref{eq:cwt} and \eqref{eq:icwt}, is typically associated with high computational costs. We, therefore, construct the wavelet space of input using the discrete wavelet transform (DWT), which minimizes both computational and memory requirements.
The DWT algorithm involves the modification of the mother wavelet to compute wavelet coefficients at scales that are typically limited to a value of $\zeta = 2^m$, where $m$ represents the decomposition level. 
The wavelet coefficients at lower decomposition levels are generally corrupted with signal noises, whereas the wavelet coefficients at higher wavelet decomposition levels contain the most important features of the input signal. Therefore, in WIB, only the wavelet coefficients of a finite number of higher-level wavelet decomposition levels are used to parameterize the kernel $\mathcal{R}_{\phi}$. 
We note that $B$, $d_v$, and $n$ represent the batch size, uplifting space, and sample length, respectively. For an uplifted input of dimension $x \in \mathbb{R}^{B \times {d_v} \times n}$, an $m$-level of DWT provides the output $\mathcal{W}\left(v_{j}\right) \in \mathbb{R}^{B \times d_{v} \times {n/2^m}}$. The uplifted dimension is integrated out by defining the kernel tensor $\mathcal{R}_{\phi} \in \mathbb{R}^{{d_v} \times d_{v} \times {n/2^m}}$. The final kernel parameterization equation is given as,
\begin{equation}\label{eq:convolution_final}
    \begin{aligned}
        \left(\mathcal{R} \cdot \mathcal{W} (v_{j})\right)_{i_b, i_o, i_n} (x) = \sum_{i_{v}=1}^{d_{v}} \mathcal{R}_{i_v, i_o, i_n} \mathcal{W} (v_{j})_{i_b, i_v, i_n}, \; {i_v, i_o \in {d_v}} .
    \end{aligned}
\end{equation}
Implementing the db6 wavelet, our system incorporates two types of WIBs within the Generator and Discriminator. The first one is the downlifting WIB that operates in the encoder part of the U-Net architecture. Second is the uplifting WIB, which is employed in the decoder part of the U-Net architecture. 
The downlifting WIB increases the lifted channel of the input function while reducing the measurement dimension. The uplifting WIB, on the other hand, shrinks the channels and restores the actual measurement length. 
The uplifting WIBs are followed by a skip connection, which concatenates the outputs of the uplifting WIBs with the outputs of corresponding downlifting WIB layers. The skip connections play a vital role in the architecture by enabling the direct flow of information from the encoder to the decoder. This facilitates the preservation of essential features and details during the transformation process.
\begin{figure}[t!]
    \centering
    \includegraphics[width=\textwidth]{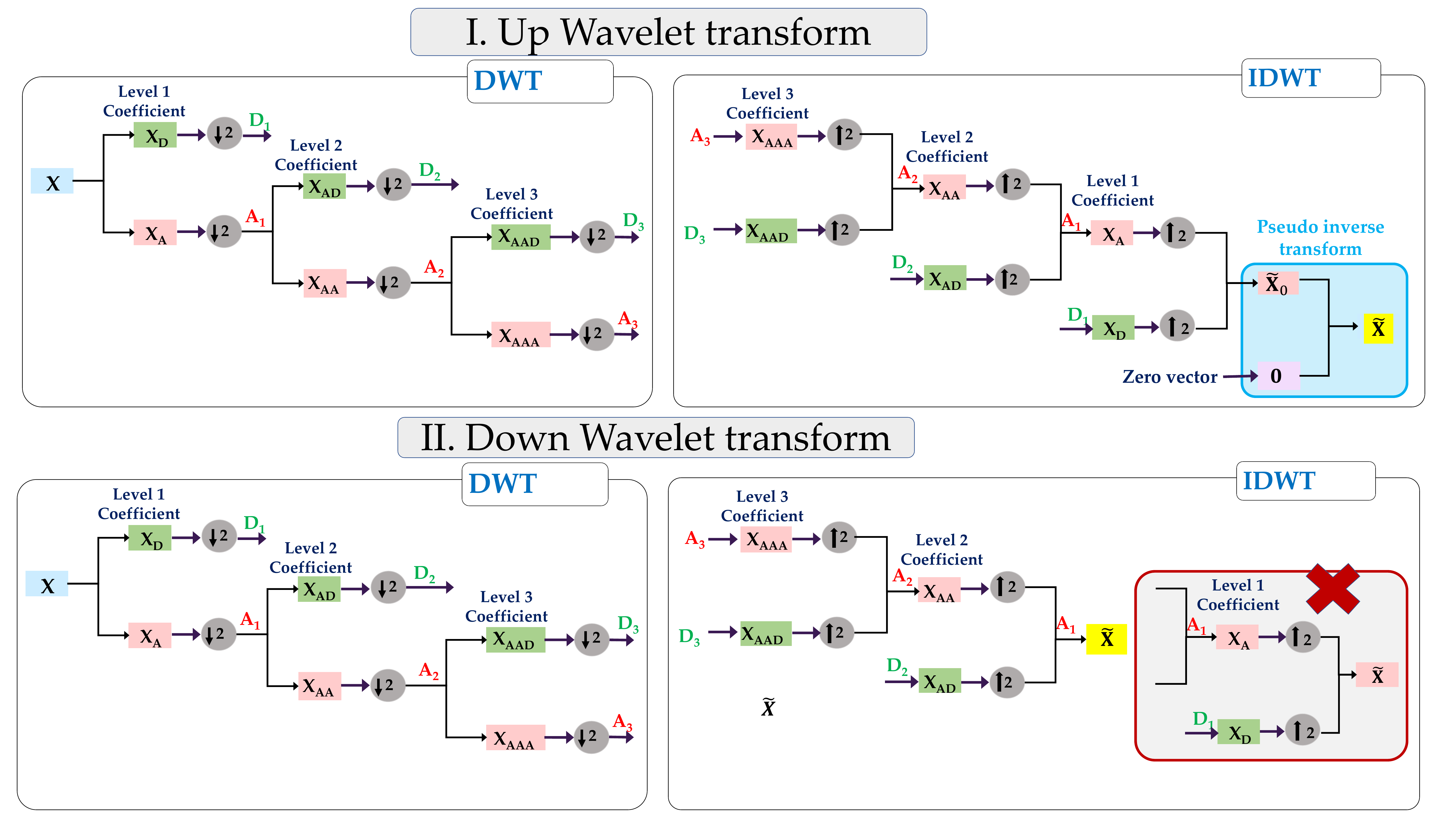}
    \caption{Schematic of uplifting and downlifting Wavelet transform}
    \label{fig:up_down_wavelet}
\end{figure}
\begin{algorithm}[ht!]
\caption{Generator Wavelet Neural Operator}
\label{algo:generator}
\begin{algorithmic}[1]
    \Require{A batch of random noises: $\bm{z}(t) \in \mathbb{R}^{B \times F \times n}$, number of recursive iterations $J$.}
    \State Apply uplifting transformation ${\rm{P}}: \mathbb{R}^{B \times F \times n} \mapsto \mathbb{R}^{B \times {C_0} \times n}$. 
    \State Perform $J$ iterations in Eq. \eqref{eq:updates} using the downlifting WIBs.
    \For{$j \gets 1$ to $J$}
        \State Perform an iteration using uplifting WIB on the output of $J^{\text{th}}$ downlifting WIB. \label{step_conc0}
        \State Concatenate the output of $(J-j)^{\text{th}}$ downlifting WIB with the output of the $j^{\text{th}}$ uplifting WIB. \label{step_upwib0}
        \State Repeat Steps \ref{step_conc0} $\to$ \ref{step_upwib0}.
    \EndFor
    \State Apply nonlinear transformation ${\rm{Q}}: \mathbb{R}^{B \times C_{o}^{\prime} \times n} \mapsto \mathbb{R}^{B \times F \times n}$. 
    \Ensure{Synthetic samples: ${\hat{\bm{x}}}(t) \in \mathbb{R}^{B \times F \times n}$.}
\end{algorithmic}
\end{algorithm}
Although both the downlifting and uplifting WIBs use the same multi-resolution representations of the parameterization space, however, the approach during inverse discrete wavelet transform (IDWT) is different among them. To understand further, let us consider the $L^{2}\left(\mathbb{R}^{n}\right)$ space, which is complete with respect to the inner product- $\langle\cdot, \cdot\rangle$. In the interest of the proposed work, we consider the wavelet and scaling basis functions,
\begin{align}\label{eq:wavelet_basis}
    \psi_{m, \tau}(x) = \frac{1}{{2^{m/2}}} \psi\left(\frac{x}{2^{m}} - \tau \right), \\
    \varphi_{m, \tau}(x) = \frac{1}{{2^{m/2}}} \varphi\left(\frac{x}{2^{m}} - \tau \right),
\end{align}
where $m \in \mathbb{Z}$ and $\tau \in \mathbb{Z}$ are the scaling and shifting parameters. 
The basis functions $\psi(\cdot)$ and $\varphi(\cdot)$ together provides the approximation space $\mathbb{A}_{m} \subset L^{2}(\mathbb{R})$ and the detail spaces $\mathbb{D}_{m} \subset L^{2}(\mathbb{R})$ at each discrete wavelet decomposition of the $m$-level multiwavelet decomposition. These subspaces contain the approximate and details wavelet coefficients of the wavelet space. These subspaces are denoted as,
\begin{equation}
    \begin{aligned}
        \mathbb{A}_{m} &= \operatorname{span}\left\{\phi_{m, \tau}\right\}_{\tau \in \mathbb{Z}}, \\ 
        \mathbb{D}_{m} &= \operatorname{span}\left\{\psi_{m, \tau}\right\}_{\tau \in \mathbb{Z}},
    \end{aligned}
\end{equation}
with $\mathbb{A}_{m} \perp \mathbb{D}_{m}$. In a standard discrete wavelet decomposition, given wavelet and scaling basis functions $\psi(\cdot)$ and $\varphi(\cdot)$, the multi-resolution analysis of the measurement length follows the property,
\begin{equation}
    \begin{aligned}
        & \{0\} \subset \mathbb{A}_{m} \subset \ldots \mathbb{A}_{h} \subset  \ldots \mathbb{A}_{1} \subset \mathbb{A}_{0} \subset L^{2}(\mathbb{R}), \\
        & \mathbb{A}_{j-1}=\mathbb{A}_{j} \oplus \mathbb{D}_{j},
    \end{aligned}
\end{equation}
where $\oplus$ denotes the orthogonal sum. However, in the cases of the proposed downlifting and uplifting WIBs, we use the downlifting and uplifting wavelet transforms (portrayed in Fig. \ref{fig:up_down_wavelet}), respectively, where the above property is modified as follows, 
\begin{align}
    \{0\} \subset \mathbb{A}_{m} \subset \ldots \subset \mathbb{A}_{h-2} \subset \mathbb{A}_{h-1} \subset L^{2}(\mathbb{R}), \label{eq:down_wib} \\
    \{0\} \subset \mathbb{A}_{m} \subset \ldots \mathbb{A}_{1} \subset \mathbb{A}_{0} \subset \{0\} \subset L^{2}(\mathbb{R}). \label{eq:up_wib}
\end{align}
Equation \eqref{eq:down_wib} corresponds to the downlifting wavelet transform, where out of a total of $m$ level DWT, the wavelet coefficients up to a level $h < m$ are neglected during IDWT. Similarly, the Eq. \eqref{eq:up_wib} corresponds to the uplifting wavelet transform, where a zero vector space is used as an extra wavelet basis to the $m$ level standard IDWT.
Using these wavelet basis functions, the wavelet decomposition $\mathcal{W}(x) \in L^{2}(\mathbb{R})$ of the input $x$ at the coarsest decomposition level $m$ can be expressed as follows,
\begin{equation}\label{eq:wavelet_expansion}
    \mathcal{W}(x) = \sum_{\tau=-\infty}^{\infty}\left\langle \mathcal{W}(x), \phi_{m, \tau}\right\rangle \phi_{m, \tau}(x)+\sum_{\tau=-\infty}^{\infty} \sum_{j=-\infty}^{m}\left\langle \mathcal{W}(x), \psi_{\tau, m}\right\rangle \psi_{\tau, m}(x).
\end{equation}
Though the above derivation can be applied to any level $h < m$, in our work, for the downlifting wavelet transform, we used $m$=2 and $h$=1. Similarly, for the uplifting wavelet transform, we have used an extra null wavelet basis for restoring the measurement dimension. For the input of size $\mathbb{R}^n$, therefore, the downlifting wavelet transform results in an output $\mathbb{R}^{n/2}$ and similarly, the uplifting wavelet transform results in an output of $\mathbb{R}^{2n}$.

\begin{algorithm}[ht!]
\caption{Discriminator Wavelet Neural Operator}
\label{algo:discriminator}
\begin{algorithmic}[1]
    \Require{Synthetic samples: ${\hat{\bm{x}}}(t) \in \mathbb{R}^{B \times F \times n}$, Actual samples: $\bm{x}(t) \in \mathbb{R}^{B \times F \times n}$, number of iterations $J$.}
    \State Concatenate generated and actual samples: $\bm{y} = \{\bm{x}, {\hat{\bm{x}}}\}$.
    \State Apply uplifting transformation ${\rm{P}}: \mathbb{R}^{B \times F \times n} \mapsto \mathbb{R}^{B \times {C_0} \times n}$ , $\bm{v} \gets {\rm{P}}(\bm{y})$.
    \State Perform $J$ iterations on $\bm{v}$ using downlifting WIBs.
    \For{$j \gets 1$ to $J$}
        \State Perform an iteration using uplifting WIB on the output of $J^{\text{th}}$ downlifting WIB. \label{step_conc}
        \State Concatenate the output of $(J-j)^{\text{th}}$ downlifting WIB with the output of the $j^{\text{th}}$ uplifting WIB. \label{step_upwib}
        \State Repeat Steps \ref{step_conc} $\to$ \ref{step_upwib}.
    \EndFor
    \State Apply nonlinear downlifting transformation ${\rm{Q}}: \mathbb{R}^{B \times C_{o}^{\prime} \times n} \mapsto \mathbb{R}^{B \times F \times n}$.
    \State Calculate the output probabilities: $\bm{r} \gets d(\bm{h}) = \int \kappa_d(t) h(t) \, dt$.
    \Ensure{Discriminating score: $\bm{r} \in \mathbb{R}^{B \times F}$.}
\end{algorithmic}
\end{algorithm}

\subsection{Fault detection and isolation employing GAWNO}
The proposed approach employs a Generative Adversarial Wavelet Neural Operator (GAWNO) model for the purpose of fault detection using reconstruction error, where the generator and discriminator components are based on wavelet neural operators. The GAWNO model is trained on a dataset comprising a 3D array format (batch, samples, features) of system variables. Through this modeling process, the GAWNO can estimate the statistical properties, specifically the mean ($\mu_s$) and standard deviation ($\sigma_s$), of reconstructed unseen data samples. The underlying hypothesis is that when the GAWNO successfully reconstructs healthy data samples, the associated prediction uncertainty remains low due to their adherence to the same underlying distribution. In contrast, the presence of faults in the reconstructed samples is expected to result in higher uncertainty. This foundational principle is leveraged for fault detection within the GAWNO framework.

The fault detection process involves two key stages. Firstly, the GAWNO is employed to reconstruct unseen data samples, and the ensuing reconstruction errors are calculated. Subsequently, prediction uncertainty is estimated by comparing these reconstruction errors to the estimated mean. If a sample's reconstruction error significantly exceeds the calculated threshold, it is flagged as potentially faulty. This initial fault detection step is followed by fault isolation, achieved by evaluating the reconstruction uncertainty of each variable independently. This isolation process entails a comparison between the reconstruction error of an observation variable and the corresponding predicted marginal posterior distribution obtained from the GAWNO model.

The essential part of this approach lies in the variable-wise assessment of reconstruction uncertainties, which facilitates the identification of variables exhibiting notable uncertainty during the data reconstruction process. These variables are considered as potential root causes underlying the detected fault, contributing to the system's anomalous behavior.

\subsection{Evaluation metrics}\label{sec:evaluation_measure}
The evaluation of the proposed method involved standard anomaly detection criteria, including metrics such as the area under the curve of the receiver operating characteristic (AUC-ROC), recall, precision, and F1-Score. The precision, recall, and F1-score are calculated as,
\begin{align}
    \mathrm{Precision} & =\frac{TP}{TP+FP}, \\
    \mathrm{Recall} &=\frac{TP}{TP+FN}, \\
    \mathrm{F}1-\mathrm{Score} &=2\times\frac{\mathrm{Precision}\times \mathrm{Recall}}{\mathrm{Precision}+\mathrm{Recall}}, 
\end{align}
where FP denotes false positive, TP denotes true positive, FN denotes false negatives, and TN denotes true negatives. For more details on the above evaluation metrics, please refer to the \cite{geiger2020tadgan}.


\section{Simulation studies}\label{sec:results}

\subsection{Hyperparameter setting}\label{sec:Gwno_architecture}
The GAWNO architecture is comprised of a generator neural operator and discriminator neural operator, where both consist of four uplifting wavelet integral blocks and four downlifting wavelet integral blocks, each activated using the GeLU activation function. The network parameters are optimized using the ADAM optimizer, with an initial learning rate of $10^{-3}$ and a weight decay factor of $10^{-5}$. The batch size during training ranges from 10 to 25, depending on the data loader's underlying mechanism. The experiments are conducted on a Quadro RTX 5000 32GB GPU. To prevent bias towards metrics with larger values, each measure is normalized using both Z-score and Min-Max scaling prior to model training. This normalization ensures that all metrics contribute equally to the training process.

\subsection{Tennessee Eastman Process (TEP) data-set}
The Tennessee Eastman process is a well-known benchmark used in the field of process control and fault detection. It was developed by the Eastman Chemical Company as a representation of a complex chemical process, and it consists of 52 variables and 21 different fault conditions, making it a rich and complex dataset for research and analysis. The Tennessee Eastman process simulates a continuous chemical plant with multiple interacting units and a wide range of operating conditions. It involves various interconnected stages, including reactors, heat exchangers, distillation columns, and separator units (the process flow diagram is illustrated in Fig. \ref{tab:my_label0}). The dataset captures the dynamic behavior and interactions between different units in the plant, making it a valuable resource for evaluating and developing advanced fault detection algorithms. The Tennessee Eastman process dataset offers a range of challenging scenarios and fault conditions, including sensor failures, valve leaks, reactor malfunctions, and other abnormal operating conditions.
The Tennessee Eastman process involves 4 gaseous reactants, A, C, D, E, and inert B, in addition to 2 liquid products, G and H, with a byproduct, F. In total, there are 41 process measurements and 12 controlled variables. This popular benchmark is comprised of 22 datasets, of which 21 (Fault 1–21) contain faults, and 1 (Fault 0) is fault-free. The details of the faults considered for testing the proposed algorithm are enlisted in Table \ref{table_0}.

\begin{table}[ht!]
    \centering
    \caption{Description of monitoring variables in the Tennessee Eastman process}
    \begin{tabular}{cl|cl}
    \toprule \textbf{ No. } & \textbf{ Variable description } & \textbf{ No. } & \textbf{ Variable description } \\
    \midrule 
    1 & A feed (stream I) & 2 & Total feed (stream 4) \\
    3 & Reactor pressure &  4 & Purge rate (stream 9) \\
    5 & Product separator pressure & 6 & Stripper pressure \\
    7 & Stripper steam flow & 8 & Separator cooling water outlet temperature \\
    9 & A feed flow valve (stream I) & 10 & Purge valve (stream 9) \\
    11 & Component A (stream 9) & 12 & Component D (stream 9) \\
    13 & D feed (stream 2) & 14 & Reactor level \\
    15 & Stripper liquid flow rate  & 16 & Stripper temperature \\
    17 & Product separator level & 18 & E feed (stream 3) \\
    19 & Product separator temperature  & 20  & Compressor work  \\
    \bottomrule
    \end{tabular}
    \label{tab:my_label0}
\end{table}
During the normal operating condition (NOC) of the Tennessee Eastman Process (TEP) simulation, the system operates in a specific production mode with a sampling period of 3 minutes. The training dataset consists of 480 samples, while the validation dataset contains 960 samples. The TEP simulation includes 21 preprogrammed faults that encompass various disturbance types and locations within the system. When a fault is introduced, the system's behavior will depend on the effectiveness of the control system in managing the disturbance. If the control system can successfully handle the fault, the system will continue to behave normally within the NOC region. However, if the control system fails to mitigate the disturbance, the system will deviate from the NOC region.
\begin{table}[ht!]
    \centering
    \caption{Description of fault types in TEP dataset. Where Fault ID 0 means Normal data while other represents the presence of fault}
    \label{table_0}
    \begin{tabular}{llll}
    \toprule
    \textbf{Behaviour} &\textbf{Fault ID} & \textbf{ Description } & \textbf{Type} \\ \midrule 
    Normal behavior & 0 & None \\ \midrule
    Back to control faults & 4  & Reactor cooling water inlet temperature & Step \\ 
    & 5 & Condenser cooling water inlet temperature & Step \\
    & 7  & C head pressure loss & Step \\ 
    \midrule 
    Controllable faults 
    & 3 & D feed temperature & Step \\& 9 & D feed temperature & Random variation \\
    & 15 & Condensor cooling water valve & Sticking \\\midrule
    Uncontrollable faults & 1 & A/C feed ratio,B composition constant & Step \\ 
    & 2  & B composition, A/C ratio constant & Step \\ 
    & 6 & A feed loss & Step         \\ 
    & 8 & A,B,C feed composition & Random variation   \\
    & 10 & C feed temperature & Random variation\\
    & 11 & Reactor cooling water inlet  & Random variation\\
    & 12 & Condenser cooling water inlet temperature & Random variation            \\ 
    & 13& Reaction kinetics & Slow Drift  \\ 
    & 14& Reactor cooling water valve & Sticking  \\ 
    & 15& Condensor cooling water valve & Sticking   \\ 
    & 16& Unknown  & Unknown \\    
    & 17& Unknown  & Unknown \\  
    & 18 & Unknown  & Unknown \\
    & 19 & Unknown  & Unknown \\
    & 20 & Unknown  & Unknown \\
    & 21 & Valve for stream 4  & Constant Position \\
    \bottomrule
    \end{tabular}
    \label{tab:my_label}
\end{table}

For each set of data corresponding to a fault condition, the simulator initially runs for 160-time points in the normal state to establish a baseline. Then, the specific fault disturbance associated with that fault condition is introduced, and the simulator continues to run for another 800 samples. This allows us to observe the system's response and behavior during the presence of the fault. This setup enables the generation of comprehensive data that includes both normal operating conditions and the effects of specific fault disturbances. It provides valuable insights into fault detection and isolation algorithms, as well as the evaluation and improvement of control systems in real-world industrial processes. The data set encompasses three distinct types of faults: controllable faults, back-to-control faults, and uncontrollable faults. Each fault type represents a different level of impact and interaction with the control system, as explained below:
\begin{enumerate}
    \item \textbf{Controllable faults }: It refers to disturbances that can be effectively compensated for by the control system. These faults do not significantly affect the process state since the control system can mitigate their impact. 
    \item \textbf{Uncontrollable faults}: It represents disturbances that cannot be effectively compensated for by the control system. These faults result in a sustained deviation from the NOC, and the control system may have limited or no ability to bring the system back to its normal state.
    \item \textbf{Back-to-control faults}: These are disturbances that initially cause the system to deviate from the normal operating condition (NOC). However, the control system is capable of compensating for at least some aspects of the disturbance over time. These faults pose a challenge as they require the control system to actively restore the system back to the NOC.
\end{enumerate}

By including these different fault types in the data set, we have evaluated and compared the performance of fault detection and isolation algorithms under various scenarios. In the case of both back-to-control and uncontrollable faults, the fault detection algorithm should ideally produce a high false detection rate (FDR) to notify the operator that the system has deviated from its original normal operating condition (NOC). In this study, all kinds of faults showing different behaviors were considered, as mentioned in Table \ref{tab:my_label}, in which fault types such as step, random variation, and slow drift are of dominant. Analyzing the effects of the different faults in the TEP dataset on 20 different variables provides valuable insights into the behavior and interdependencies of the industrial process.
We have added the different behavior faults in different variables to check the efficiency of our proposed algorithm. As mentioned below, we have provided a general overview of how each fault impacts the variables in the TEP dataset:
\begin{itemize}
    \item \textbf{Fault 1}: It affects the A/C feed ratio in Stream 4, leading to changes in the C and A feeds, causing variations in feed A in the recycle Stream 5 and subsequent adjustments in the A feed flow in Stream 1. This fault influences variables associated with material balances, such as composition and pressure.
     \item \textbf{Fault 2}: It involves a significant disturbance in the D feed flow rate, affecting variables related to material balances and potentially causing changes in composition, pressure, and flow rates.
    \item \textbf{Fault 3}: It results in variations in the reactor feed flow rate, leading to changes in reactor-related variables, such as reactor temperatures and pressures.
    \item \textbf{Fault 4}: It affects the B feed temperature, influencing variables related to temperature, including reactor temperatures and temperatures in associated streams.
    \item \textbf{Fault 5}: It affects the D feed temperature, leading to changes in variables linked to temperature, such as reactor temperatures and temperatures in related streams.
    \item \textbf{Fault 6}: It involves a disturbance in the E feed temperature, influencing variables related to temperature, such as reactor temperatures and temperatures in associated streams.
    \item \textbf{Fault 7}: It results in variations in the E feed flow rate, affecting material balances and potentially leading to changes in composition, pressure, and flow rates.
    \item \textbf{Fault 8}: It affects the condenser cooling water flow rate, leading to variations in cooling-related variables, including temperatures and flows in cooling water streams.
    \item \textbf{Fault 9}: It results in a disturbance in the cooling water flow rate to the reactor feed effluent exchanger, influencing variables tied to heat exchange, such as temperatures and flows in heat exchanger streams.
    \item \textbf{Fault 10}: It affects the heat exchanger's effectiveness, causing variations in heat exchanger-related variables, including temperatures, pressures, and heat exchange rates.
    \item \textbf{Fault 11}: It involves a disturbance in the reactor coolant temperature, impacting variables related to temperature, such as reactor temperatures and temperatures in associated streams.
    \item \textbf{Fault 12}: It affects the cooling water temperature to the condenser, leading to variations in cooling-related variables, including temperatures and flows in cooling water streams.
    \item \textbf{Fault 13}: It results in a disturbance in the cooling water temperature to the reactor feed effluent exchanger, impacting variables tied to heat exchange, such as temperatures and flows in heat exchanger streams.
    \item \textbf{Fault 14}: It affects the cooling water temperature to the reactor intercooler, causing variations in cooling-related variables, including temperatures and flows in cooling water streams.
    \item \textbf{Fault 15}: It involves a disturbance in the reactor coolant flow rate, influencing variables related to reactor operation, such as temperatures and pressures.
    \item \textbf{Fault 16}: It affects the reactor feed flow rate to the separator, leading to variations in separator-related variables, such as flow rates, compositions, and pressures.
    \begin{figure}[b!]
    \centering
    \includegraphics[width=\textwidth]{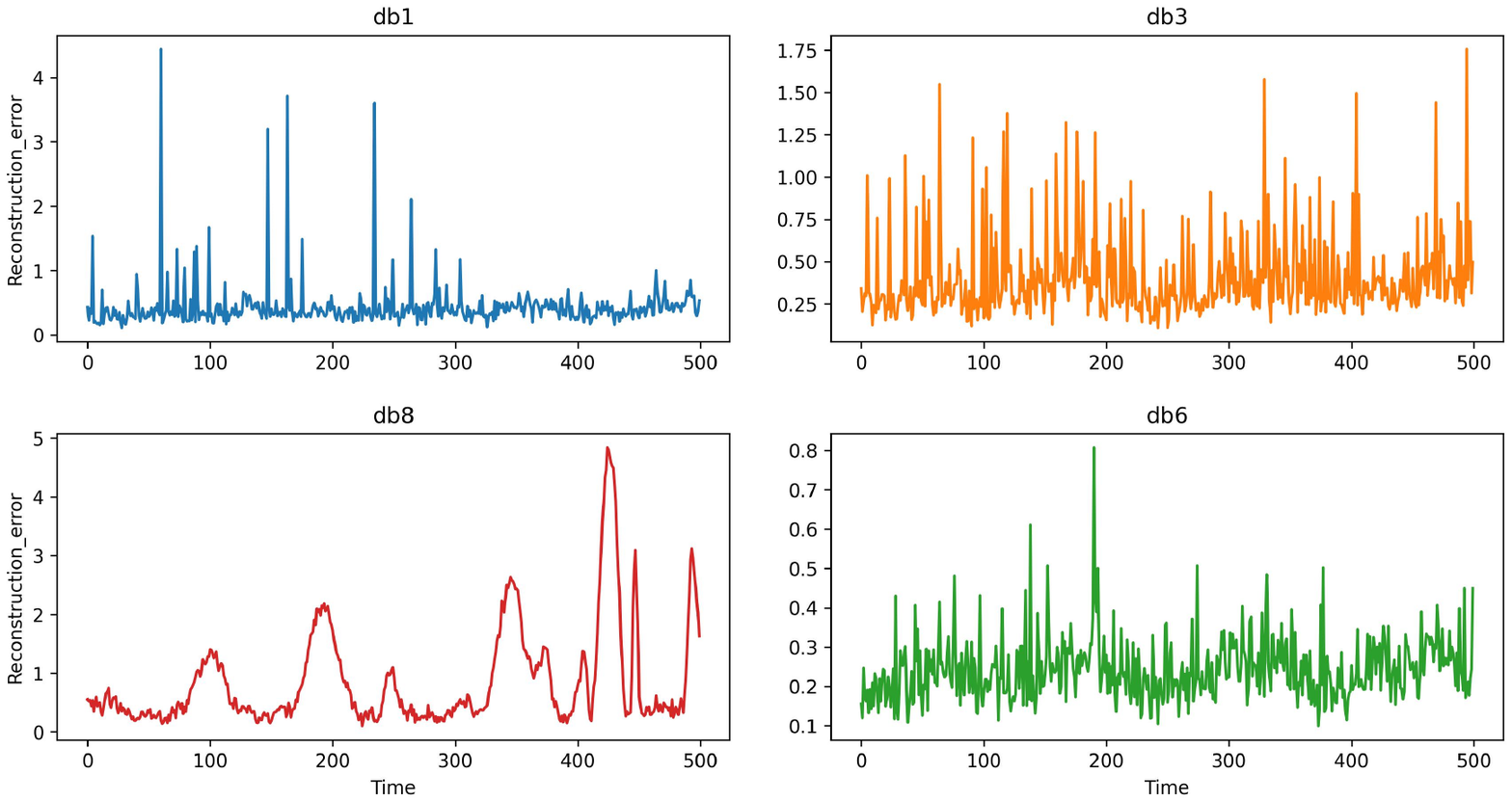}
    \caption{Average reconstruction error plot during training on the Tennessee Eastman Process dataset using GAWNO. The training procedure incorporated different wavelets (db1, db3, db6, and db8). Notably, db6 demonstrated superior learning of the dataset distribution, yielding a significantly low reconstruction error.}
    \label{fig:rec_ten}
    \end{figure}
    \item \textbf{Fault 17}: It results in a disturbance in the separator level control, impacting variables related to the operation of the separator and subsequent process units.
    \item \textbf{Fault 18}: It involves a step change in the reactor coolant flow rate, influencing variables related to coolant flow rates and reactor temperatures.
    \item \textbf{Fault 19}: It causes a disturbance in the reactor inlet temperature, affecting variables associated with reactor temperatures and potentially influencing other related variables.
    \item \textbf{Fault 20}: It results in variations in the vapor fraction controller in Stream 4, impacting vapor fractions and potentially causing changes in compositions and flow rates in related streams.
     \item \textbf{Fault 21}: It involves a disturbance in the pressure controller in Stream 4, affecting variables related to stream pressure and potentially causing variations in flow rates and compositions.
\end{itemize}

\begin{figure}[!ht]
     \centering
     \begin{subfigure}[b]{\textwidth}
         \centering
        \includegraphics[width=0.88\textwidth]{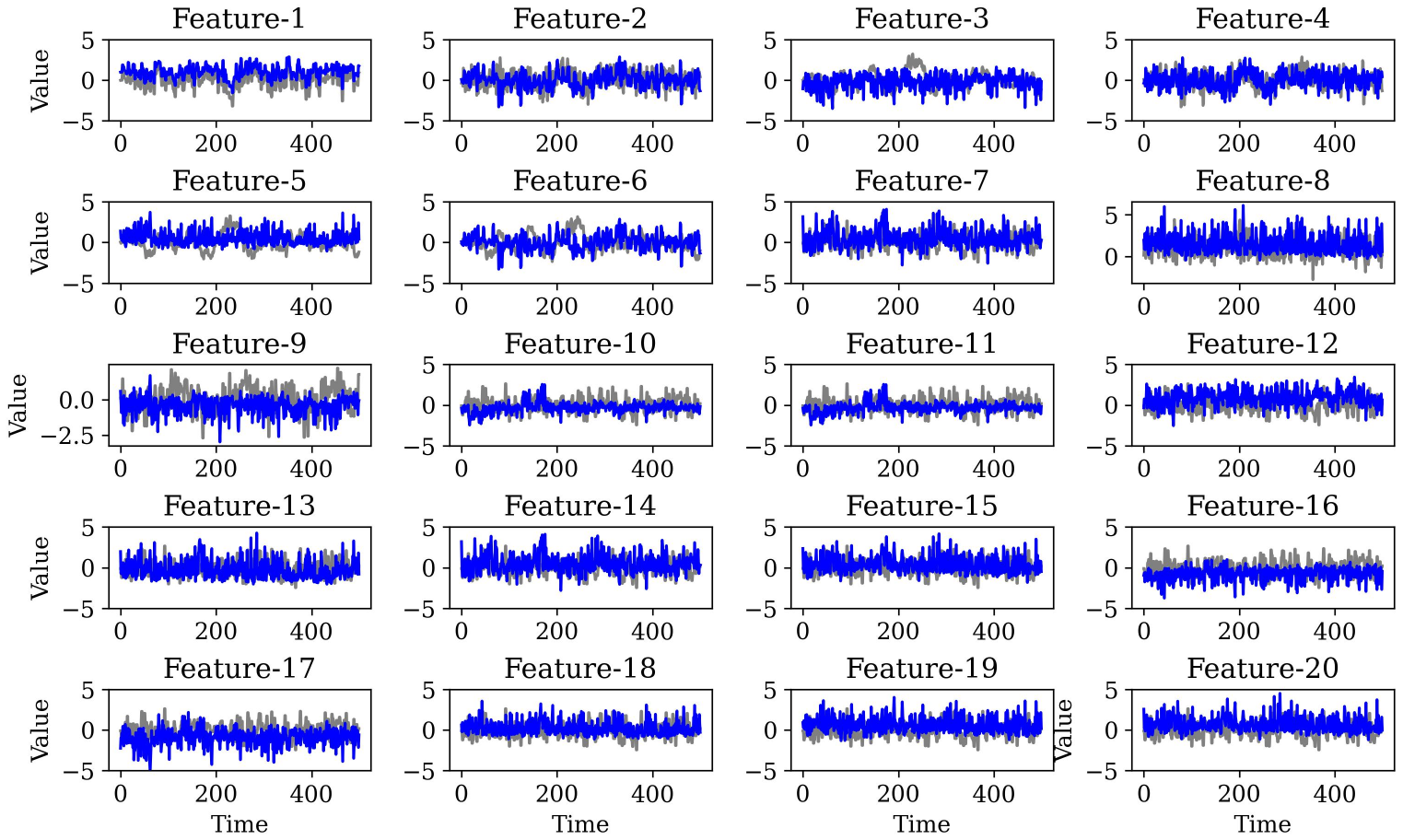}

        \caption{Training with Tennessee Eastman Process dataset}
        \label{fig:tcmm2}
     \end{subfigure}
     \hfill
     \begin{subfigure}[b]{\textwidth}
         \centering
         \includegraphics[width=0.88\textwidth]{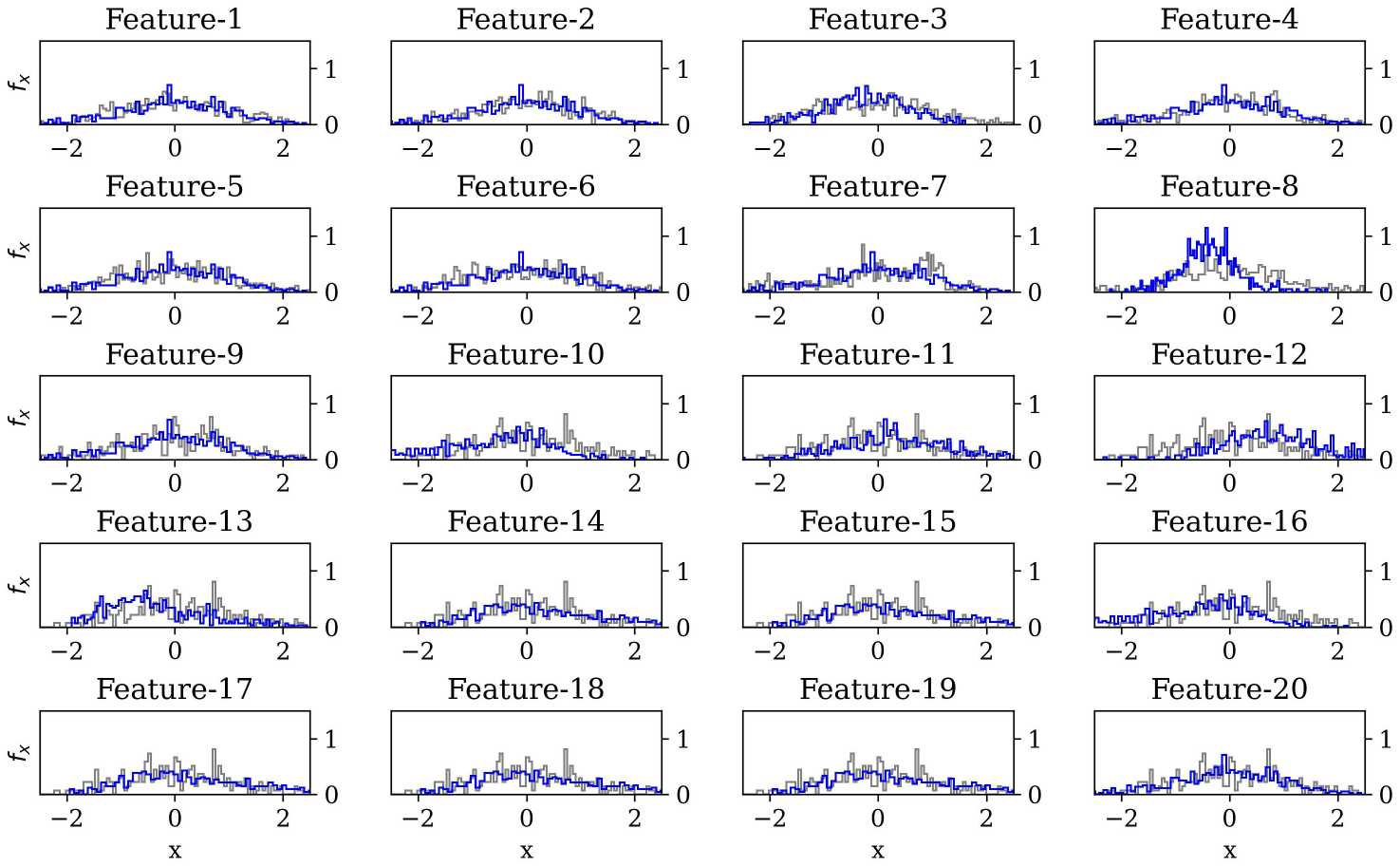}
        \caption{Distribution learning by GAWNO}
        \label{fig:tcmm3}
     \end{subfigure}
        \caption{Training Results using GAWNO for Tennessee Eastman Process dataset. The input function sample comprises random noise, while the WNO-based generator generates the data. The generator is trained to learn the underlying data distribution. (a) Training with Tennessee Eastman Process dataset where grey represents actual data while blue represents the generated data (b) Data distribution histogram resulting from the GAWNO approach.}
        \label{fig:tcall9}
\end{figure}

\subsubsection{Fault detection and isolation in Tennessee Eastman Process (TEP) data-set}
\begin{figure}
    \centering
    \includegraphics[width=0.5\textwidth]{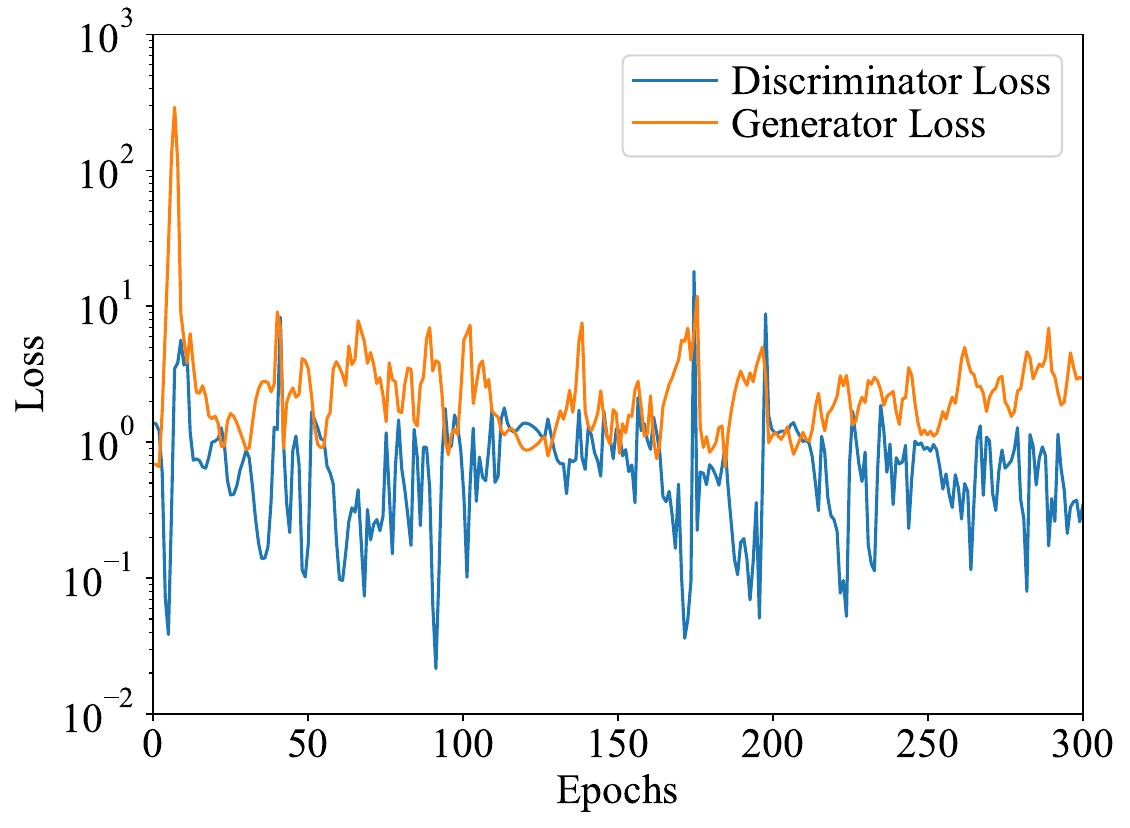}
    \caption{Training Result Analysis using GAWNO on the Tennessee Eastman Process dataset. The training procedure involves two integral components: the Discriminator and the Generator. The training loss of the Discriminator, representing its ability to differentiate real data from generated data, exhibits a gradual decline throughout the training epochs. Concurrently, the Generator's training loss, which signifies its capacity to produce data indistinguishable from genuine data, also experiences a consistent reduction.}
    \label{fig:loss_tennessee}
\end{figure}

The TEP dataset consists of a multitude of variables representing process measurements and operational parameters. These variables are interdependent and exhibit complex dynamics, making fault detection and isolation a challenging task. In the context of fault detection and isolation, GAWNO can be leveraged to learn the normal operating patterns of the TE process and generate synthetic data samples that capture the underlying distribution. The generator network is trained on normal operating data, while the discriminator network learns to distinguish between real and generated samples. Once the GAWNO is trained, the generator network can be used to reconstruct the input data, generating a synthetic version of the original data. The reconstruction error analysis is a key component in fault detection using GANs. 
\begin{figure}[ht!]
    \centering
    \includegraphics[width=\textwidth]{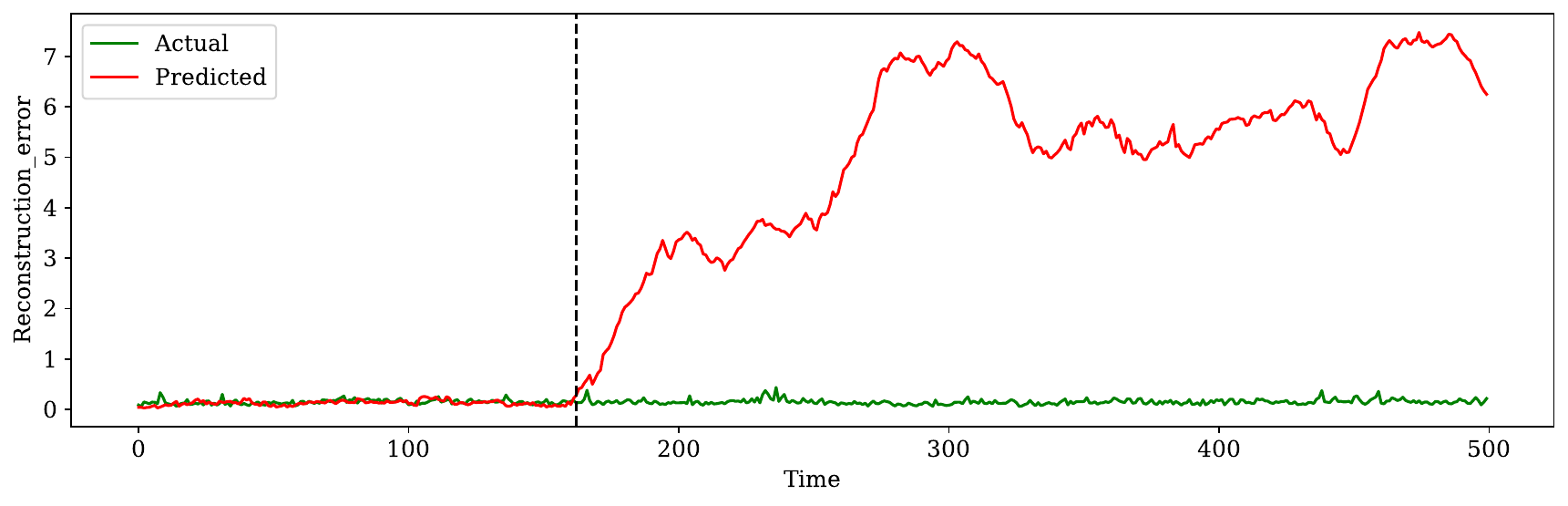}
    \caption{Testing average reconstruction error plot using GAWNO on the Tennessee Eastman Process dataset  }
    \label{fig:overall_rec}
\end{figure}
\begin{figure}[ht!]
    \centering
    \includegraphics[width=0.9\textwidth]{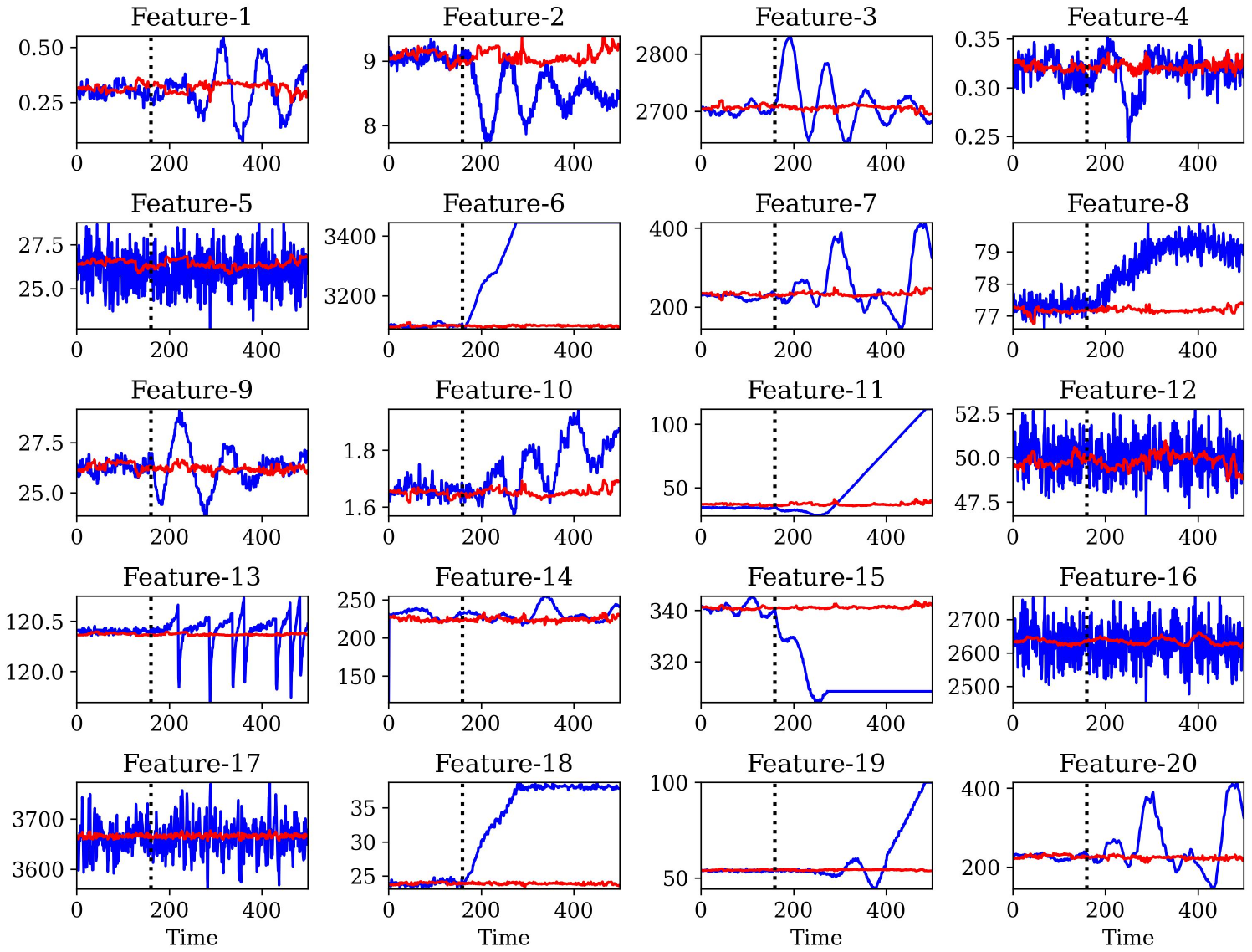}
    \caption{Testing Results using GAWNO on the Tennessee Eastman Process dataset: Following the training phase, where the generator was trained to learn the data distribution, this figure showcases the testing outcomes. By introducing faults at time step 160 in each variable during testing, we aimed to evaluate the model's fault detection and reconstruction capabilities. This observation emphasizes the GAWNO effective learning of the data distribution, as it struggles to conform to deviations from the learned distribution}
    \label{fig:test}
\end{figure}
\begin{figure}[ht!]
    \centering
     \includegraphics[width=0.9\textwidth]{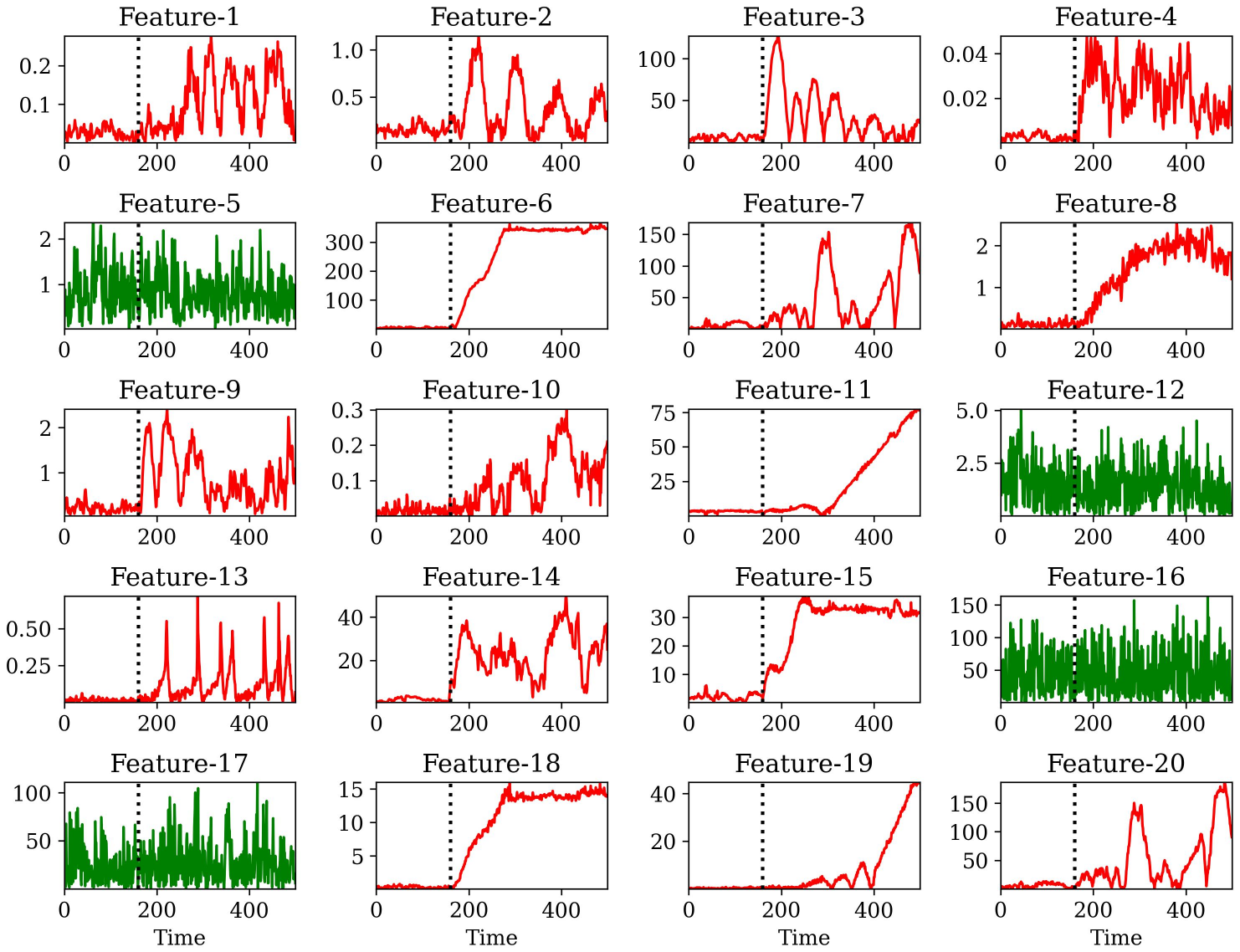}
     
    \caption{Reconstruction Error Analysis using GAWNO on the Tennessee Eastman Process dataset: This figure presents the results of reconstruction error analysis conducted during testing. The generator was trained to comprehend the data distribution, and fault introduction was carried out at time step 160 in select variables during testing. Notably, the model exhibited challenges in accurately reconstructing the introduced faults, resulting in elevated reconstruction errors, as depicted by the RED plot. In contrast, the GREEN plot demonstrates low reconstruction error when dealing with normal data, indicating successful fault absence detection within the data.}
    \label{fig:reconstruct}
\end{figure}
It quantifies the dissimilarity between the original data and its reconstructed counterpart. By calculating the reconstruction error for each data point, deviations from normal behavior can be identified. During training, we generated an average reconstruction error plot that involved various wavelets, including db1, db3, db6, and db8. Notably, the training process revealed that db6-based wavelet effectively learned the dataset distribution, resulting in a remarkably low reconstruction error, as shown in Fig. \ref{fig:rec_ten}. While during testing, higher reconstruction errors indicate potential anomalies or faults in the process. A threshold-based approach is typically employed to determine whether the reconstruction error exceeds an acceptable limit. Anomalies and potential faults are detected by comparing the reconstruction error with a predefined threshold. During training, the GAN gradually learns to generate realistic samples that conform to the distribution of the normal data by analyzing and visualizing the distribution of reconstruction errors through a histogram plot as shown in Fig. \ref{fig:tcmm3}. The learning process is also portrayed in Fig. \ref{fig:loss_tennessee}. 

GAWNO offers a unique advantage in capturing the complex dependencies and distributions within the data as shown in Fig. \ref{fig:tcmm2}, allowing for the identification of specific variables or components contributing to abnormal behavior. Fig. \ref{fig:overall_rec} showcases the average reconstruction error plot for normal data and faulty data for the TEP dataset. As observed,  multivariate data seems to have an anomalous pattern as the reconstruction error plot shows spikes because of the uncertainty in the reconstruction after the 160th time step. Once a fault is detected, the next goal is to identify the main variables associated with the fault. Without using labeled fault examples, this step involves determining the observation variables with the abnormal deviations, which are most relevant to locating and troubleshooting the faults. To determine which variables deviate abnormally, each observation variable is compared to its corresponding predicted marginal posterior distribution estimated from the samples. 
The fault isolation process using GAWNO involves analyzing the reconstruction errors for each variable and setting appropriate thresholds to determine the significance of the deviations. Variables with higher reconstruction errors indicate a stronger contribution to the fault occurrence. By employing statistical techniques or domain knowledge, it is possible to determine the critical variables responsible for the fault and isolate them from the rest of the system. Experimental evaluations on the TEP dataset have demonstrated the effectiveness of GAWNO in fault isolation for multivariate variable systems as shown in Fig. \ref{fig:test} after 160-time steps predicted sample shows different behaviour compared to normal one because of the presence of different behavior faults. This confirms the presence of faults in the system at different variables.  Thus, this approach helps in identifying the variable that causes the anomalous behavior by showing high reconstruction errors as depicted in Fig. \ref{fig:reconstruct} with RED plots, whereas normal data with low reconstruction error is represented by GREEN plot. The fault detection performance results which involved various wavelets, including db1, db3, db6, and db8 are presented in Table \ref{table_dbcom} in Appendix, with the overall best performance highlighted in bold.

\subsection{$\mathrm{WWTP} \mathrm{N}_2 \mathrm{O}$ dataset}
The dataset contains a complete set of measurements corresponding to the Avedore wastewater treatment plant (WWTP) and $\mathrm{N}_2 \mathrm{O}$ emissions, making it a significant resource to use when developing effective $\mathrm{N}_2 \mathrm{O}$ emission management techniques \cite{ahn2010n2o,chen2019assessment}. The dataset includes measurements of
$\mathrm{NH}_4{ }^{+}-\mathrm{N}, \mathrm{NO}_3{ }^{-}-\mathrm{N}$, dissolved oxygen $(\mathrm{DO}$), and water temperature which are constantly monitored in both compartments of all four reactors, while liquid-phase $\mathrm{N}_2 \mathrm{O}$ is additionally measured using Unisense $\mathrm{N}_2 \mathrm{O}$ sensors in two reactors (i.e., Reactor 1 and Reactor 3). The measuring sensors are placed in the same location with respect to the water flow direction in each compartment thus, we limited our study to Reactor 3 data. The influent flow rate into each reactor is also recorded online using a flow meter. Composite samples over $24 \mathrm{~h}$ are analyzed every 1 or 2 weeks to indicate influent quality.
\begin{figure}[ht!]
    \centering
    \includegraphics[width=0.7\textwidth]{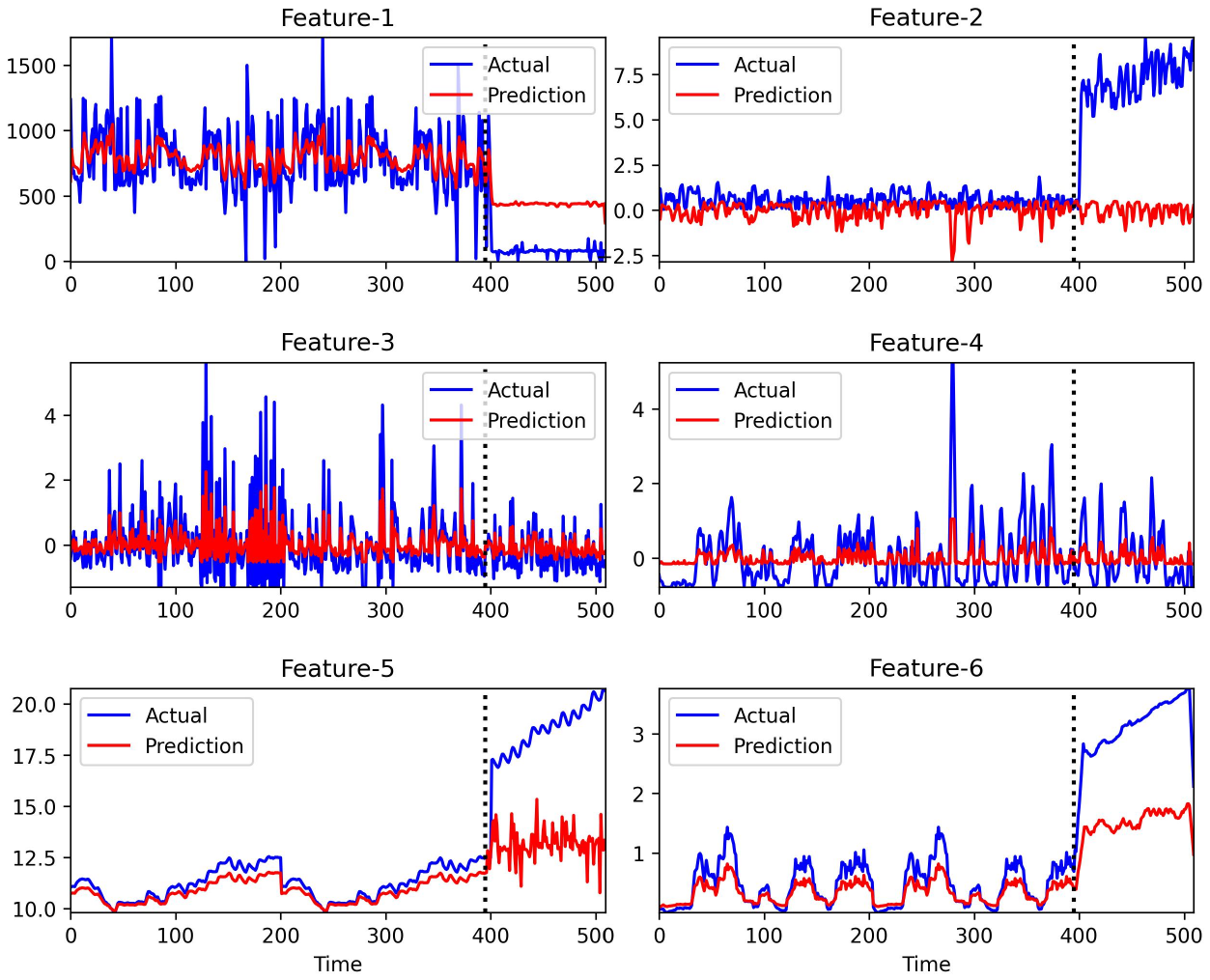}
    \caption{Testing Results using GAWNO on Avedore Wastewater Treatment Plant Data: After training the generator to comprehend the data distribution, this figure illustrates the outcomes of the testing phase. Introducing faults at time step 395 for each variable during testing aimed to assess the model's fault detection and reconstruction capabilities. However, the model exhibited challenges in accurately reconstructing the introduced faults and adhering to the expected distribution. This observation shows the model's ability to handle deviations from the learned distribution within the Avedore wastewater treatment plant data.}
    \label{fig:water_test}
\end{figure}

\subsubsection{Fault detection and isolation in $\mathrm{WWTP} \mathrm{N}_2 \mathrm{O}$ dataset}
Similar to the TEP dataset, the $\mathrm{WWTP} \mathrm{N}_2 \mathrm{O}$ dataset comprises a multitude of variables representing various aspects of wastewater treatment and $\mathrm{N_2O}$ emissions (schematic diagram of Carrousel reactor is shown in Table \ref{fig:water} in Appendix). The intricate relationships among these variables create challenges in identifying faults within the system. The GAWNO framework can be employed to address these challenges by learning the normal operating patterns of the wastewater treatment process and generating synthetic data samples that capture the underlying distribution.
During the training phase of the GAWNO model, the generator network learns to generate synthetic data points that closely resemble the distribution of normal data, while the discriminator network is trained to distinguish between real data and the generated samples. This adversarial training process equips the model to differentiate between normal and faulty conditions in the $\mathrm{WWTP} \mathrm{N}_2 \mathrm{O}$ dataset. Throughout the training phase, the GAWNO model gradually learns to produce realistic samples that align with the normal data distribution as depicted in Fig. \ref{fig:train_water} in the Appendix.
Figure \ref{fig:loss_water} in the Appendix also details the training loss profile for both the discriminator and generator networks along with the reconstruction error in Fig. \ref{fig:test_waterdata} and Fig. \ref{fig:reconstruct_water}, contributing to a comprehensive understanding of the model's learning progression. One critical aspect of fault detection using GAWNO is the reconstruction error analysis. Higher reconstruction errors are indicative of potential anomalies or faults in the wastewater treatment process. A threshold-based approach, similar to the one used in the TEP dataset, can be employed to determine whether the reconstruction error exceeds an acceptable limit, thus indicating the presence of faults or anomalies.
Empirical assessments performed on the $\mathrm{WWTP} \mathrm{N}_2 \mathrm{O}$ dataset underscore the effectiveness of the GAWNO approach in fault isolation. Notably, in Fig. \ref{fig:water_test}, anomalies in the $\mathrm{WWTP} \mathrm{N}_2 \mathrm{O}$ emissions can be identified through the deviation of predicted samples from normal patterns after 160-time steps. This divergence in behavior is indicative of the presence of different behavior faults within the $\mathrm{WWTP} \mathrm{N}_2 \mathrm{O}$ dataset. With the help of the reconstruction error plot, it was observed that high reconstruction errors indicate the presence of fault denoted by the RED plot. Whereas, the GREEN plot represents low reconstruction error when dealing with normal data (Fig. S6). In conclusion, the GAWNO approach holds promise for fault detection and isolation in the $\mathrm{WWTP} \mathrm{N}_2 \mathrm{O}$ dataset.

\subsection{Performance comparison of GAWNO with baseline methods}
The fault detection performance of GAWNO was compared with other widely used models, including Principal Components Analysis (PCA), Dynamic Principal Components Analysis (DPCA), Long Short-Term Memory (LSTM), and Autoencoders (AE), and GANs. Evaluation measures from Section \ref{sec:evaluation_measure} were used to assess and rank the performance of the proposed approach.
\begin{table}[h!]
\centering
 \begin{subtable}{\textwidth}
     \centering
     \begin{tabular}{|c|c|c|c|c|c|c|}
        \hline Methods & Recall & Precision & F1-Score & AUC & FP &  FN \\
        \hline
        GAN      &  0.789       & 0.838   & 0.813  & 0.795 & 18 & 20\\
        \hline
        LSTM      &    0.8668    & 0.939   &  0.901  & 0.888 & 8 & 13\\
        \hline
        LSTM-AE & 0.9318 & 0.989 & 0.9590  & 0.949 & 6 & 8\\
        \hline
        DPCA & 0.987 & 0.69 & 0.812 & 0.815 & 31 & 7\\
        \hline
        PCA & 0.808 & 0.675 & 0.735 & 0.71 & 32 & 16\\
        \hline
        \textbf{GAWNO} & \textbf{0.9910} & \textbf{0.996}  & \textbf{0.993}    & \textbf{0.981} & \textbf{3} & \textbf{2} \\
        \hline
     \end{tabular}
    \caption{Average performance metrics for $\mathrm{WWTP} \mathrm{N}_2 \mathrm{O}$ data set in presence of fault}
    \label{table_a}
     \vspace{5mm}
 \end{subtable}
\\
 \begin{subtable}{\textwidth}
     \centering
     \begin{tabular}{|c|c|c|c|c|c|c|}
        \hline Methods & Recall & Precision & F1-Score & AUC & FP & FN \\
        \hline GAN     &  0.74     &  0.795   &  0.767 & 0.753 & 12 & 17 \\
        \hline LSTM      &  0.812      &  0.88  &  0.845  & 0.839 & 10 & 22 \\
        \hline
        LSTM-AE & 0.871  & 0.922 &   0.8953     & 0.889 & 7 & 16 \\
        \hline
        DPCA & 0.979 & 0.59 & 0.736 & 0.718 & 26 & 8\\
        \hline
        PCA & 0.76 & 0.51 & 0.618 & 0.604 & 28 & 19\\
        \hline
        \textbf{GAWNO} & \textbf{0.989} & \textbf{0.99}  & \textbf{0.989}    & \textbf{0.980} & \textbf{4} & \textbf{5} \\
        \hline
     \end{tabular}
    \caption{Average performance metrics for Tennessee Eastman  data set in presence of fault}
     \label{table_bb}
     \vspace{3mm}
 \end{subtable}
 \caption{Average performance metrics compared with baselines }
 \label{table_results}
\end{table}
\begin{enumerate}
    \item  Among the compared models, PCA and DPCA are widely recognized techniques. For the $\mathrm{WWTP} \mathrm{N}_2 \mathrm{O}$ dataset, PCA reduced the dimensionality of the data from 6 to 2 dimensions. In the case of DPCA, the RR-13 algorithm \cite{rato2013defining} was employed with a lag estimation of 2. To capture approximately 99$\%$ of the variance, 4 principal components (PCs) were selected. Similarly, for the TEP dataset, the original data was mapped to 6 dimensions using the PCA method, while for DPCA, a lag of 1 and 14 PCs was chosen to capture around 99$\%$ of the variance.
    \item Next, LSTM was employed, using a neural network architecture with three LSTM layers of 30, 40, and 50 units, respectively. Each layer was followed by a dense layer with a single unit that forecasted the value at the subsequent time step.
    \item As for the Autoencoder (AE) approach, we employed standard autoencoders with dense layers. The decoder and encoder sections of the dense autoencoder consisted of three dense layers each. For the $\mathrm{WWTP} \mathrm{N}_2 \mathrm{O}$ dataset, the encoder layers had 6, 18, and 36 units, respectively, in each layer, while the decoder layers had 36, 18, and 6 units, respectively, in each layer. For the TEP dataset, the encoder architecture included layers with 16, 64, and 256 units, respectively, while the decoder architecture featured layers with 256, 64, and 16 units.
\end{enumerate}
The fault detection performance results are presented in Table \ref{table_results}, with the overall best performance highlighted in bold. Upon comparing various algorithms, GAWNO stood out as the top-performing method. Specifically, for the  $\mathrm{WWTP} \mathrm{N}_2 \mathrm{O}$ dataset (see \ref{table_a}), GAWNO achieved the highest scores for AUC (0.981) and recall (0.991), showcasing its superior ability to accurately detect faults while minimizing false negatives. Moreover, GAWNO demonstrated a significantly low number of false positives, with only 4 instances reported. Similar positive trends were observed for the TEP dataset, where GAWNO outperformed AE, LSTMs, and other algorithms for fault detection Table \ref{table_bb}. The statistical analysis in Table \ref{table_results} further confirmed GAWNO's outstanding performance for fault detection across different time steps, including starting, intermediate, and endpoint faults, in both the TEP and  $\mathrm{WWTP} \mathrm{N}_2 \mathrm{O}$ datasets. Regarding accuracy and F1-Score, GAWNO surpassed other popular methods such as GANs, AE, LSTM, DPCA, and PCA. Overall, the proposed approach has the potential to enhance safety, process control, and maintenance, leading to more sustainable operations.


\section{Conclusion}\label{sec:conclusion}
A noble generative adversarial wavelet neural operator (GAWNO) is proposed for learning probability distributions of underlying processes. It integrates the concepts of existing generative adversarial networks (GAN) and a recently proposed wavelet neural operator (WNO), where the WNO architecture is utilized as the generator and discriminator within a U-Net framework. 
The use of the WNO as the generator in the GAWNO allows for the generation of multivariate synthetic data that closely resembles the healthy multivariate data distribution. On the other hand, the use of WNO as a discriminator helps in correctly distinguishing between real and synthetic data, directing the generator to capture the underlying probability distributions of the healthy data accurately. The U-Net architecture also plays a crucial role in enhancing the performance of fault detection. With its skip connections, the U-Net facilitates the integration of multi-scale features and enables the effective extraction of relevant information from the input data. This enables the network to capture both local and global patterns, leading to improved fault detection accuracy. As an application, the reconstruction error between the real and synthetic data is utilized as a measure for fault detection and isolation in industrial processes. By leveraging the power of the WNO and generative networks, the proposed approach demonstrates higher accuracy, sensitivity, and specificity in detecting faults in multivariate variables, which is further proof of the effectiveness of the proposed GAWNO framework.

\section*{Acknowledgements} 
T. Tripura acknowledges the financial support received from the Ministry of Education (MoE), India, in the form of the Prime Minister's Research Fellowship (PMRF). S. Chakraborty acknowledges the financial support received from Science and Engineering Research Board (SERB) via grant no. SRG/2021/000467, and from the Ministry of Port and Shipping via letter no. ST-14011/74/MT (356529).

\section*{Declarations}


\subsection*{Conflicts of interest} The authors declare that they have no conflict of interest.

\subsection*{Availability of data and material} Upon acceptance, all the data to reproduce the results in this study will be made available to the public on GitHub by the corresponding author.

\subsection*{Code availability} Upon acceptance, all the source codes to reproduce the results in this study will be made available to the public on GitHub by the corresponding author.

\appendix

\section{Appendix}

\subsection{Process flow diagram of Tennessee Eastman Process (TEP)}\label{appen:flow_tennessee}
The Tennessee Eastman process simulates a continuous chemical plant with multiple interacting units and a wide range of operating conditions. It involves various interconnected stages, including reactors, heat exchangers, distillation columns, and separator units. The process flow diagram is provided below. The description of the units is given in Table \ref{tab:my_label0}.
\begin{figure}[ht!]
    \centering
    \includegraphics[width=0.95\textwidth]{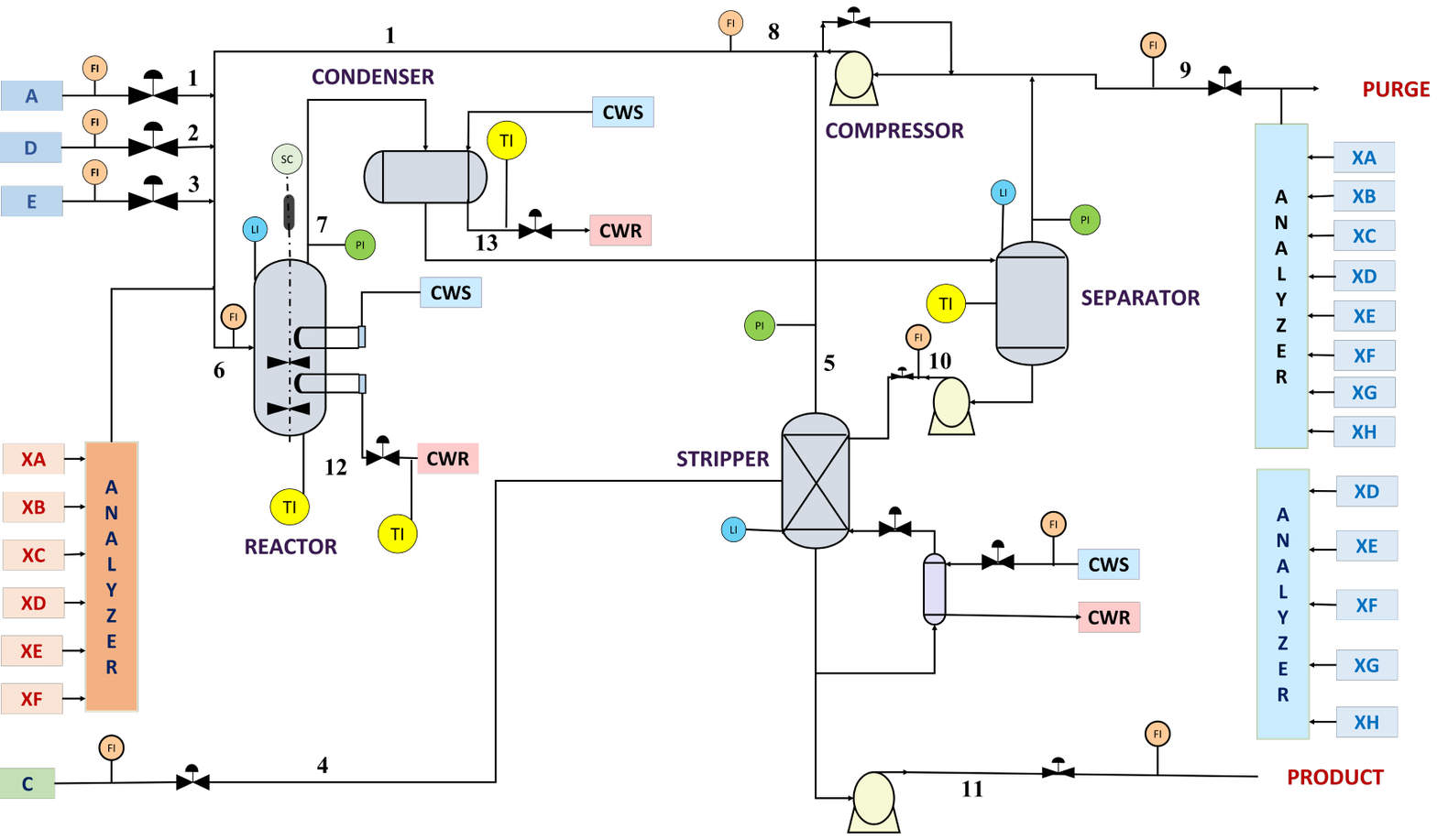}
    \caption{Process flow diagram of Tennessee Eastman process}
    \label{fig:ten}
\end{figure}

\subsection{Effect of different wavelets on fault detection performance}
The fault detection performance results against various wavelets, including db1, db3, db6, and db8, are presented in Table \ref{table_dbcom}, with the overall best performance highlighted in bold.
\begin{table}[h!]
\centering
    \begin{tabular}{|c|c|c|c|c|c|c|}
        \hline Methods & Recall & Precision & F1-Score & AUC \\
        \hline
        db1 & 0.93 & 0.91 & 0.919 & 0.90 \\
        \hline 
        db3 & 0.96 & 0.98 & 0.969 & 0.95 \\
        \hline
        \textbf{db6} & \textbf{0.989} & \textbf{0.99}  & \textbf{0.989} & \textbf{0.980}  \\
        \hline
        db8 & 0.88  & 0.92 &   0.89     & 0.889 \\
        \hline
    \end{tabular}
    \caption{Average performance metrics for Tennessee Eastman  data set in the presence of fault incorporating different wavelets (db1, db3, db6, and db8)}
    \label{table_dbcom}
\end{table}

\subsection{Training with Avedore wastewater treatment plant data}
\begin{figure}[hb!]
    \centering
    \includegraphics[width=0.55\textwidth]{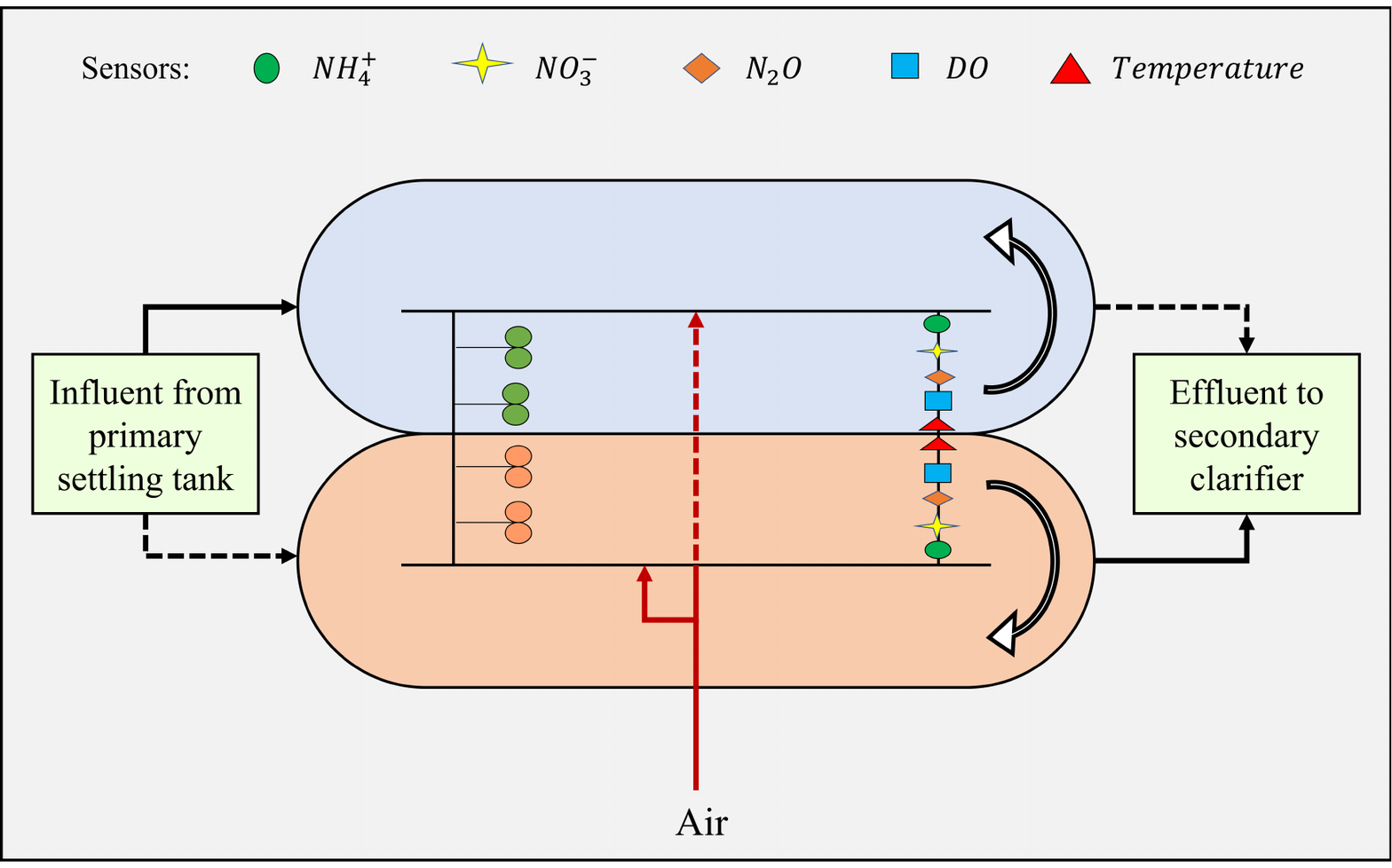}
    \caption{Schematic diagram of Carrousel reactor comprising two alternatively fed and intermittently aerated compartments (Locations of measuring sensors with respect to water flow direction are marked)}
    \label{fig:water}
\end{figure}

\begin{figure}[ht!]
    \centering
    \includegraphics[width=0.95\textwidth]{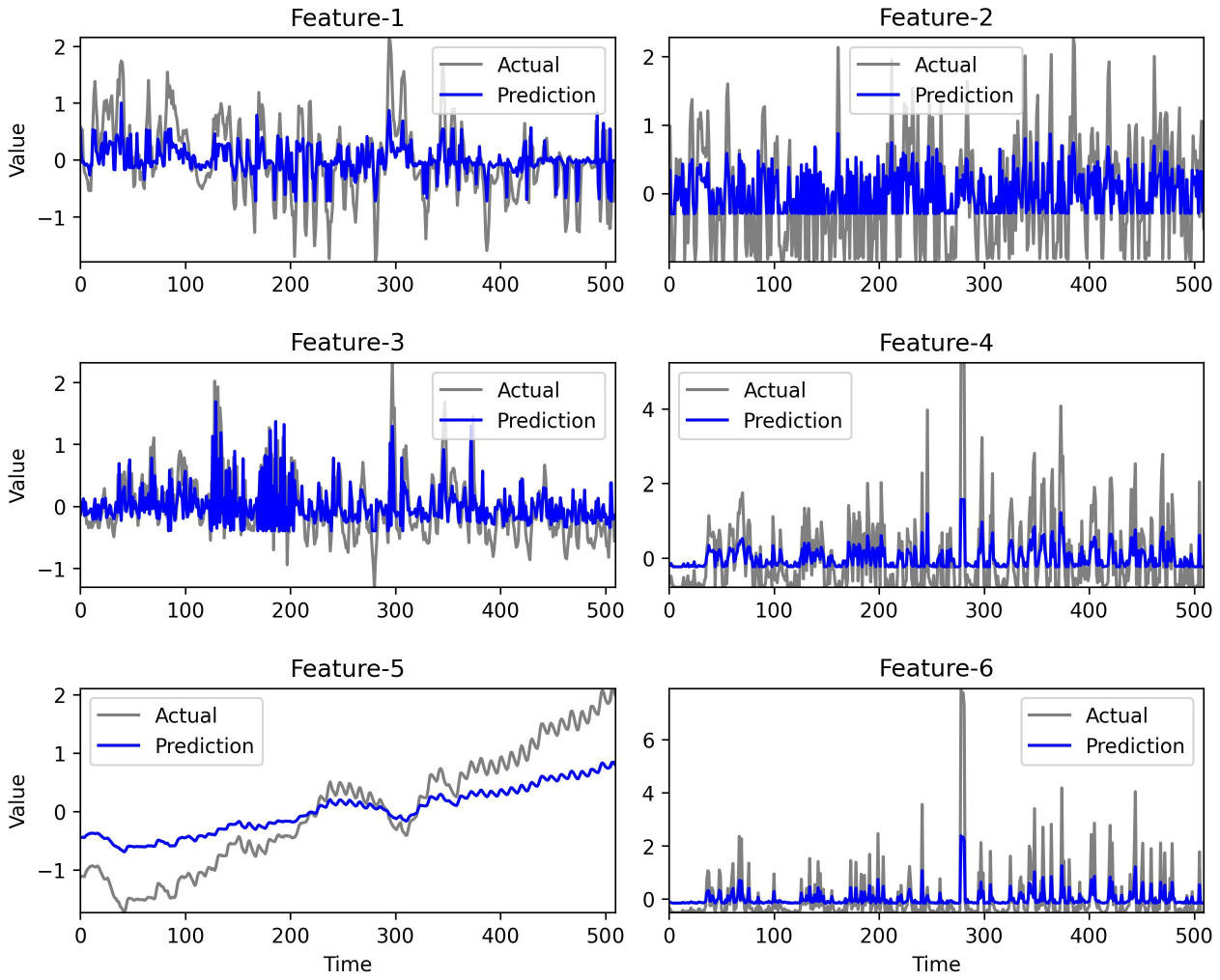}
    \caption{Training with Avedore wastewater treatment plant data }
    \label{fig:train_water}
\end{figure}

\begin{figure}[ht!]
    \centering
    \includegraphics[width=0.5\textwidth]{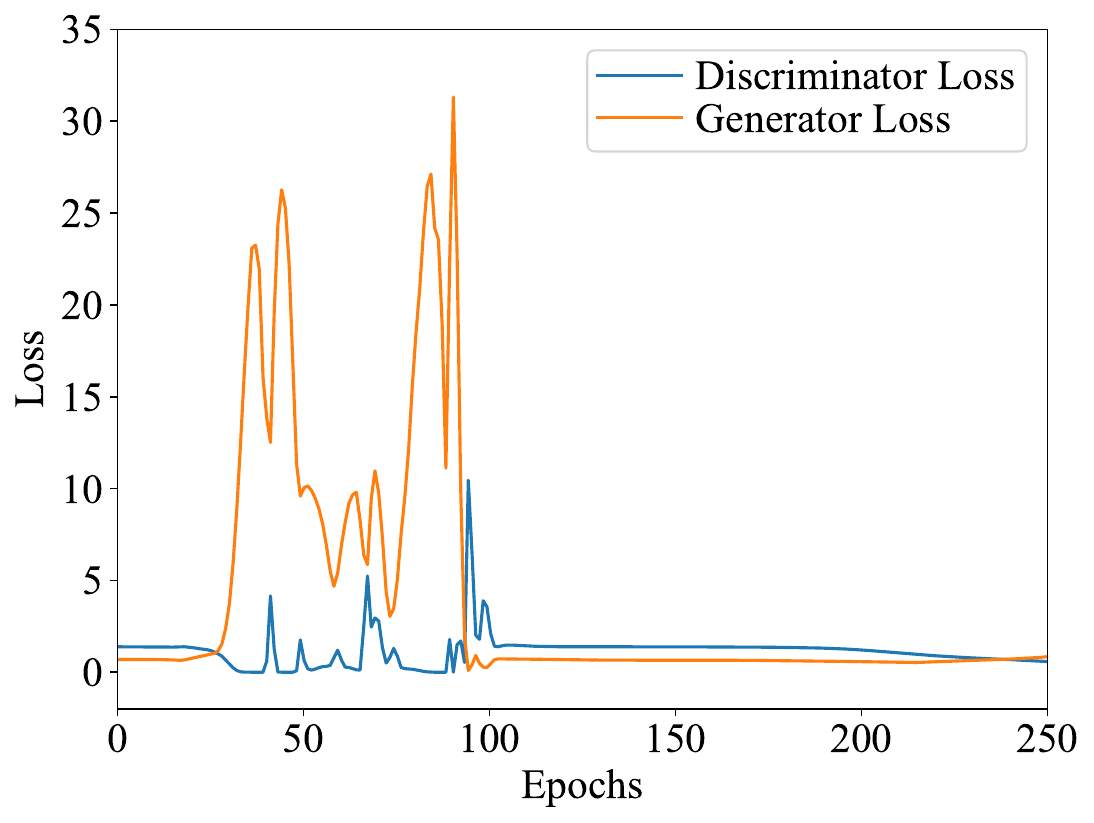}
    \caption{Discriminator and Generator training loss for Avedore wastewater treatment plant dataset visualization using GAWNO. The interaction between the Discriminator and Generator leads to a convergence of their respective training losses. This convergence signifies the model's progressive improvement in generating data that effectively captures the intricate data distribution within the Avedore wastewater treatment plant dataset.}
    \label{fig:loss_water}
\end{figure}

\begin{figure}[ht!]
    \centering
    \includegraphics[width=\textwidth]{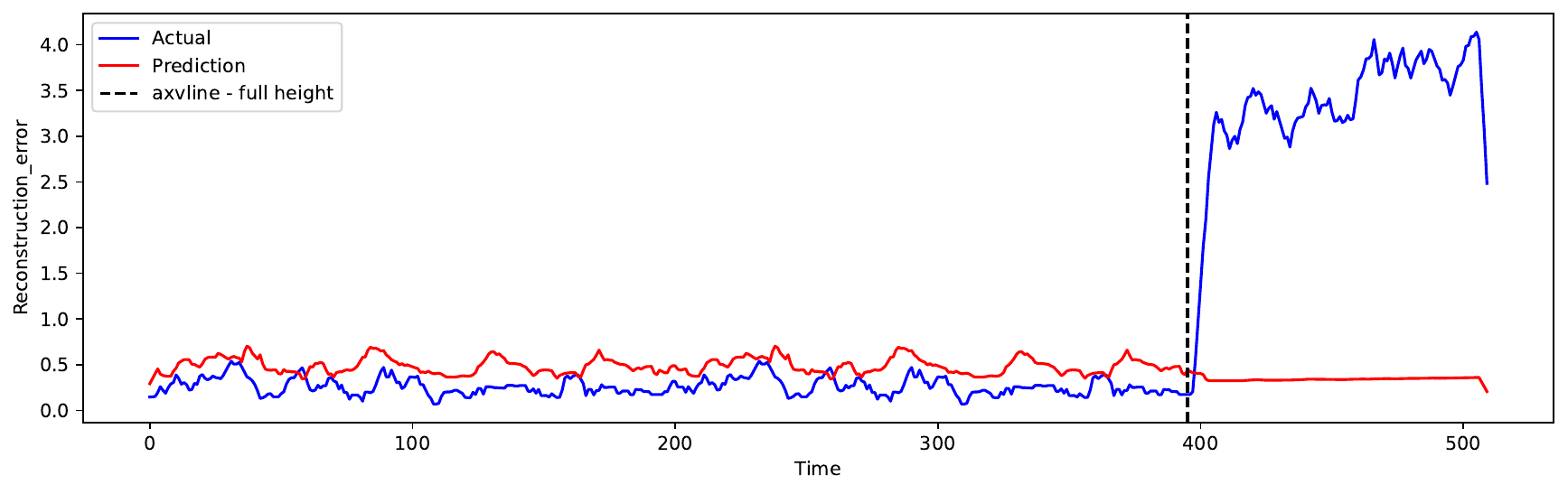}
    \caption{Overall reconstruction error with Avedore wastewater treatment plant data}
    \label{fig:test_waterdata}
\end{figure}

\begin{figure}[ht!]
    \centering
    \includegraphics[width=0.9\textwidth]{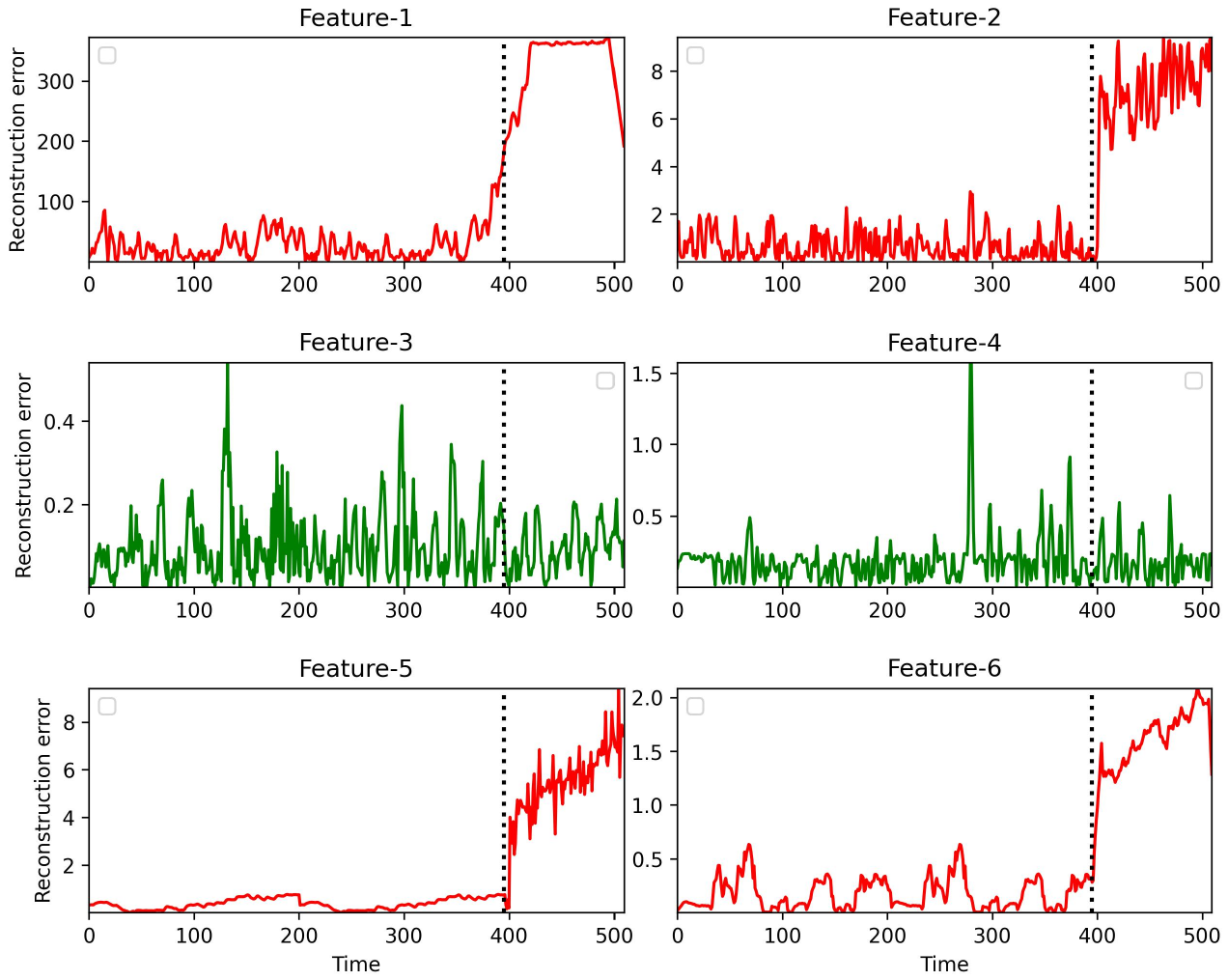}
    \caption{Reconstruction error plot of Avedore wastewater treatment plant data. GAWNO exhibited challenges in accurately reconstructing the faults, resulting in high reconstruction errors denoted by the RED plot. Whereas the GREEN plot represents low reconstruction error when dealing with normal data, indicating successful fault absence detection within the data.}
    \label{fig:reconstruct_water}
\end{figure}

Avedore wastewater treatment plant (WWTP) and it's $\mathrm{N}_2 \mathrm{O}$ emissions, rendering it a valuable resource for the development of effective $\mathrm{N}_2 \mathrm{O}$ emission management strategies. The dataset incorporates recordings of key parameters such as $\mathrm{NH}_4{ }^{+}-\mathrm{N}$, $\mathrm{NO}_3{ }^{-}-\mathrm{N}$, dissolved oxygen $(\mathrm{DO}$), and water temperature. We limited our study to Reactor 3 data (the schematic diagram of the reactor is depicted in Fig. \ref{fig:water}). During the training phase, the GAWNO model learns to synthesize data that captures the normal data distribution, simultaneously discerning between typical and faulty conditions. Fig.\ref{fig:train_water} illustrates the training with Avedore wastewater treatment plant data, whereas Fig. \ref{fig:loss_water} represents the discriminator and generator training loss. Furthermore, the model's proficiency in reconstructing faulty data plays a pivotal role in identifying specific components or stages within the wastewater treatment process responsible for $\mathrm{N}_2 \mathrm{O}$ emissions, as illustrated in Fig. \ref{fig:test_waterdata}. By assessing reconstruction errors between the generated and actual data, the GAWNO model effectively identifies anomalous patterns indicative of faults or irregularities within the WWTPN2O dataset, as depicted in the accompanying Fig. \ref{fig:reconstruct_water}.



\end{document}